\documentclass[10pt,twocolumn,letterpaper]{article}

\usepackage{cvpr}
\usepackage{makecell}
\usepackage{pifont}
\usepackage{csquotes}
\usepackage{times}
\usepackage{epsfig}
\usepackage{graphicx}
\usepackage{amsmath}
\usepackage{amssymb}
\usepackage{mathtools}
\usepackage{multicol}
\usepackage{caption}
\usepackage{float}
\usepackage{subcaption}
\usepackage[pagebackref=true,breaklinks=true,letterpaper=true,colorlinks,bookmarks=false]{hyperref}

\newcommand\blfootnote[1]{%
  \begingroup
  \renewcommand\thefootnote{}\footnote{#1}%
  \addtocounter{footnote}{-1}%
  \endgroup
}

\cvprfinalcopy % *** Uncomment this line for the final submission

\ifcvprfinal\fi
\begin{document}

%%%%%%%%% TITLE
\title{Unsupervised Part-Based Disentangling of Object Shape and Appearance}

\author{Dominik Lorenz \hspace{1cm} Leonard Bereska \hspace{1cm} Timo Milbich \hspace{1cm} Bj\"orn Ommer \\
Heidelberg Collaboratory for Image Processing / IWR, Heidelberg University
}

\maketitle
\thispagestyle{empty}

\begin{abstract}
Large intra-class variation is the result of changes in multiple object characteristics. Images, however, only show the superposition of different variable factors such as appearance or shape. Therefore, learning to disentangle and represent these different characteristics poses a great challenge, especially in the unsupervised case. Moreover, large object articulation calls for a flexible part-based model. We present an unsupervised approach
for disentangling appearance and shape by learning parts consistently over all instances of a category. Our 
model for learning an object representation 
is trained by simultaneously exploiting invariance and equivariance constraints between synthetically transformed images. Since no part annotation or prior information on
an object class is required, the approach is applicable to arbitrary classes.
We evaluate our approach on a wide range of object categories and diverse tasks including pose prediction, disentangled image synthesis, and video-to-video translation. The approach outperforms the state-of-the-art on unsupervised keypoint prediction and compares favorably even against supervised approaches on the task of shape and appearance transfer.
\end{abstract}

\section{Introduction}
\begin{figure}
	\centering
	\includegraphics[trim={4.3cm 0.5cm 6.2cm 0cm},clip, width=1.\linewidth]{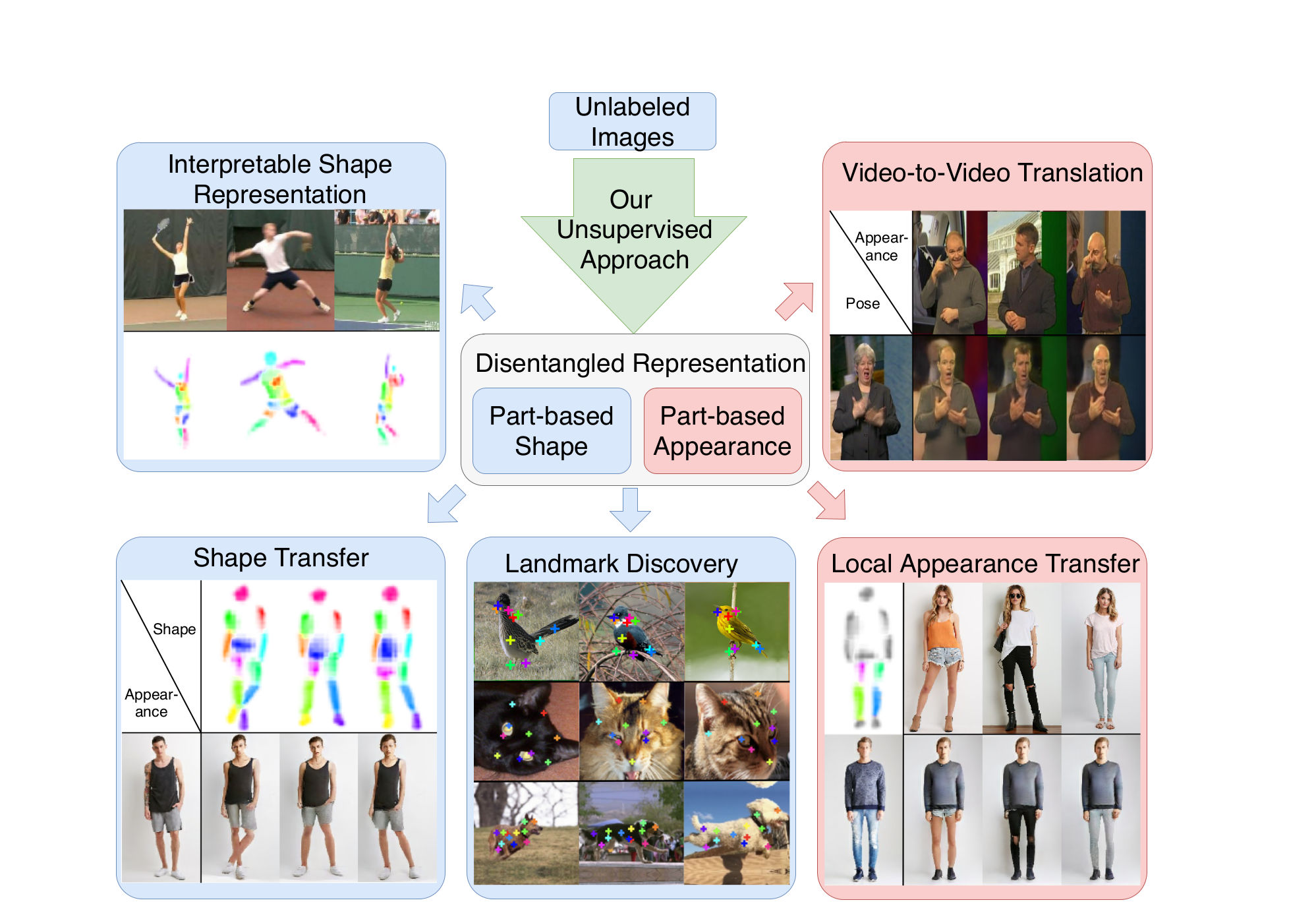}
	\caption{Our unsupervised learning of a disentangled part-based shape and appearance enables numerous tasks ranging from unsupervised pose estimation to image synthesis and retargeting. For more results visit the project page \protect\footnotemark}
	\label{fig:firstpage}
\end{figure}
\footnotetext{https://compvis.github.io/unsupervised-disentangling/}

A grand goal of computer vision is to automatically, without supervision information, learn about the characteristics of an object in the world.  Typically, images show the interplay of multiple such factors of variation. We want to disentangle \cite{Desjardins2012dr, Bengio2013rep, Chen2016infogan, Higgins2016betavae, Eastwood2018dr} the effects of these different characteristics and imagine, i.e., synthesize, new images where they are altered individually. For instance, after observing a number of different unlabeled instances of an object category, we want to learn their variations in shape (such as pose relative to the viewer and body articulation) and appearance, \eg, texture and color differences in fur/clothing or skin color. Disentangling shape and appearance is particularly challenging because object deformation typically leads to complicated
\enquote{recoloring} of image pixels \cite{Shu:2018ua,Esser:2018ue}: moving a limb may change the color of former background pixels into foreground and vice versa.

To address the disentangling problem for shape and appearance, several supervised methods have been proposed lately~\cite{Ma:2017wq, Ma:2017uu, deBem:2018wp, Esser:2018ue, Siarohin:2018wk, Balakrishnan:2018wo}. By conditioning generative models on a pre-specified shape representation, they are able to successfully explain away appearance. However, they are limited to object categories, for which pose labels are readily available such as human bodies and faces, but they cannot be applied to the vast amounts of unlabelled data of arbitrary object classes.

For unsupervised learning, instead of taking a known shape to capture all non-shape factors, both shape and appearance need to be learned  simultaneously.
Recently some unsupervised approaches have been proposed to disentangle these factors~\cite{Shu:2018ua, Xing:2018un}. However, these works have only shown results for rather rigid objects, like human faces or require several instances of the same person \cite{Denton:2017uf}.

Object variation can be global, such as difference in viewpoint, but it is oftentimes local (animal tilting its head, person with/without jacket), thus calling for a local, disentangled object representation.
The traditional answer are compositions of rigid parts~\cite{Fischler1973rep, Fergus2003object, Felzenszwalb:2010ve}.
In the context of recent unsupervised shape learning an instantiation
of this are landmarks~\cite{Thewlis:2017wi, Zhang:2018vz, Jakab:2018wc}.
In this paper, we propose the first approach to learn a part-based disentangled representation of shape \textit{and} appearance for articulated object classes 
without supervision and from scratch.
In the spirit of analysis-by-synthesis~\cite{Yildirim:2015ur}, we learn the factors by a generative process.
We formulate
explicit equivariance and invariance constraints an object representation should fulfill
and incorporate them in a fully differentiable autoencoding framework.

Our approach yields significant improvements upon the state-of-the-art in unsupervised object shape learning, evaluated on the task of landmark regression.
We compare to competitors on a wide range of diverse datasets both for rigid and articulated objects, with particularly large gains for strong articulations.
Furthermore, our disentangled representation of shape and appearance competes favorably even against state-of-the-art supervised results.
We also show disentangling results on the task of video-to-video translation, where fine-grained articulation is smoothly and consistently translated on a frame-to-frame level.
Lastly, since our representation captures appearance locally, it is also possible to transfer
appearance on the level of individual object parts. An overview of the scope of possible applications is given in Fig.~\ref{fig:firstpage}.

\section{Related Work}
\textbf{Disentangling shape and appearance.}
	Factorizing an object representation into shape and appearance is a popular ansatz for representation learning.
	Recently, a lot of progress has been made in this direction by conditioning generative models on shape information \cite{Esser:2018ue, Ma:2017wq, deBem:2018wp, Ma:2017uu, Siarohin:2018wk, Balakrishnan:2018wo}.
	While most of them explain the object holistically, only few also introduce a factorization into parts \cite{Siarohin:2018wk, Balakrishnan:2018wo}.
	In contrast to these shape-supervised approaches, we learn both shape and appearance without any supervision.
	
	For unsupervised disentangling, several generative frameworks have been proposed ~\cite{Higgins2016betavae, Chen2016infogan, Li2018analogy, Denton:2017uf, Shu:2018ua, Xing:2018un}.
	However, these works use holistic models and show results on rather rigid objects and simple datasets, while we explicitly tackle strong articulation with a part-based formulation.

\textbf{Part-based representation learning.}
	Describing an object as an assembly of parts is a classical paradigm for learning an object representation in computer vision \cite{Ross:2006uc, Nguyen:2013vk, Cootes:1998tn, eigenstetter:CVPR:2014}. 
	What constitutes a part, is the defining question in this scheme. Defining parts by visual and semantic features
	or by geometric shape and its behavior under viewpoint changes and object articulation in general leads to a different partition of the object. Recently, part learning has been mostly employed for discriminative tasks, such as in \cite{Felzenszwalb:2010ve, Novotny:2017ta, Singh:2012un, Mesnil:2013hi, Yang:2016uo, Lam:2017ta}.
	To solve a discriminative task, parts will encode their semantic connection to the object and can ignore the spatial arrangement and articulation. In contrast, our method is driven by an image modelling task.
	Hence, parts have to encode both spatial structure \textit{and} visual appearance accurately.

\textbf{Landmark learning.}
	There is an extensive literature on landmarks as compact representations of object structure.
	Most approaches, however, make use of manual landmark annotations as supervision signal \cite{Wu:2017vc, Ranjan:2016vv, Yu:2016vi, Zhang:2016vx, Zhu:2015tz, Zhang:2014wy, ufer2017deep, Pedersoli:2014ta, Ionescu:2011ue, Toshev:2014tp, Pfister:2015uo, Wei:2016ws, Newell:2016vq, Lim:2018uo, Cao:2017vv}.

	To tackle the problem without supervision, Thewlis \etal \cite{Thewlis:2017wi} proposed enforcing equivariance of landmark locations under artificial transformations of images. The equivariance idea had been formulated in earlier work \cite{Lenc:2016tz} and has since been extended to learn a dense object-centric coordinate frame \cite{Thewlis:2017wg}. However, enforcing only equivariance encourages consistent landmarks at 
	discriminable object locations,
	but disregards an explanatory coverage of the object.

	Zhang \etal \cite{Zhang:2018vz} addresses this issue: the equivariance task is supplemented by a reconstruction task in an autoencoder framework, which gives visual meaning to the landmarks. However, in contrast to our work, he does not disentangle shape and appearance of the object. Furthermore, his approach relies on a separation constraint in order to avoid the collapse of landmarks.
	This constraint results in an artificial, rather grid-like layout of landmarks, that does not scale to complex articulations.
	
	Jakab \etal \cite{Jakab:2018wc} proposes conditioning the generation on a landmark representation from another image. A global feature representation of one image is combined with the landmark positions of another image to reconstruct the latter. Instead of considering landmarks which only form a representation for spatial object structure, we factorize an object into local parts, each with its own shape \textit{and} appearance description.
	Thus, parts are learned which meaningfully capture the variance of an object class in shape as well as in appearance.

	Additionally, and in contrast to all these works (\cite{Thewlis:2017wi, Zhang:2018vz, Jakab:2018wc}) we take the extend of parts into account, when formulating our equivariance constraint. Furthermore, we explicitly address the goal of disentangling shape and appearance on a part-based level by introducing invariance constraints.

\begin{figure*}[ht!]
    \centering
    \includegraphics[trim={2cm 3cm 7cm 7cm},clip, width=0.87\linewidth]{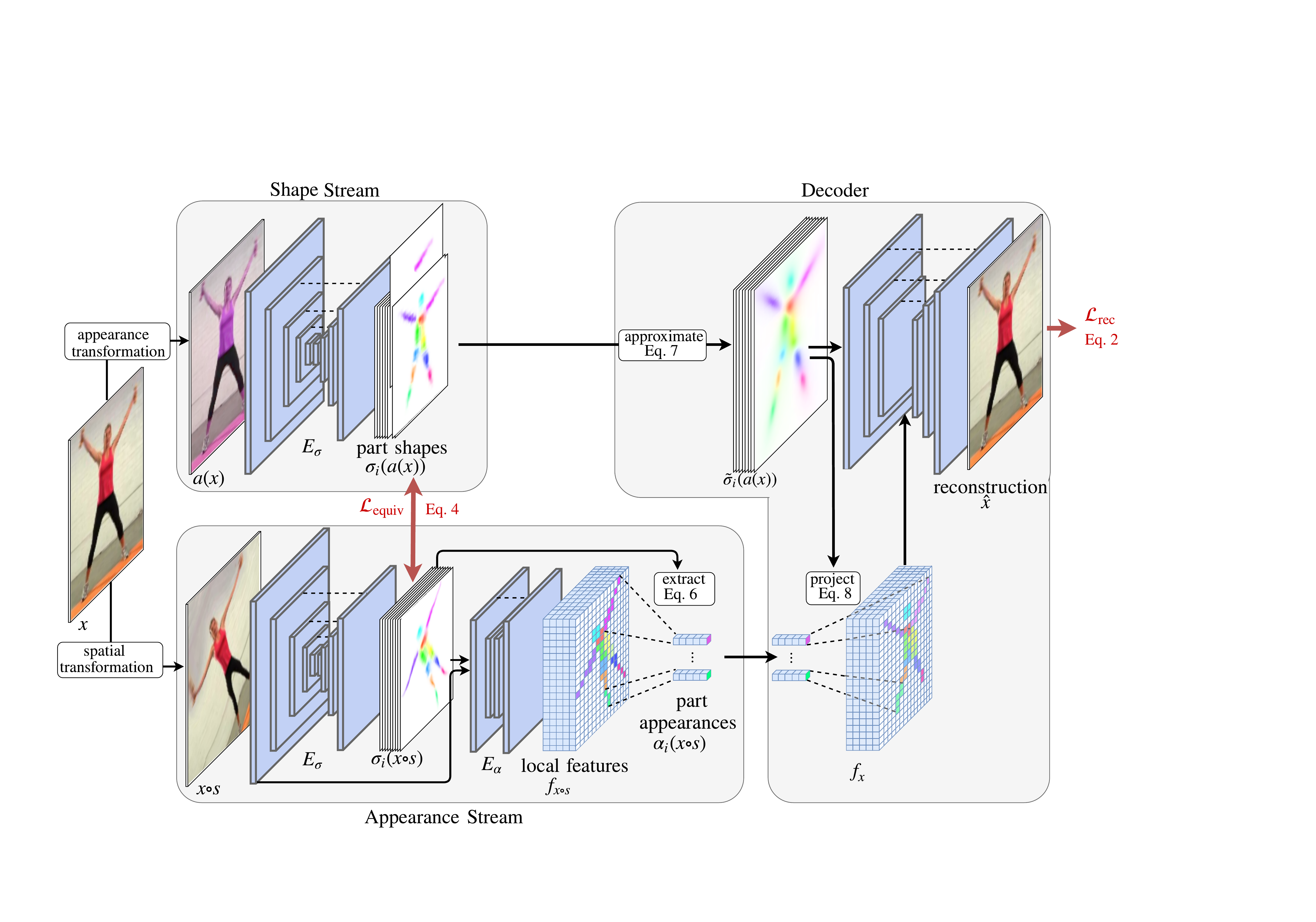}
	\caption{Two-stream autoencoding architecture for unsupervised learning of object shape and appearance.} 
    \label{fig:architecture}
\end{figure*}
\section{Approach}
    Let $x: \Lambda \rightarrow \mathbb{R}$ be an image portraying an object and background clutter. $\Lambda \subset \mathbb{N}^2$ is the space of image coordinates. Now consider an image $x': \Lambda \rightarrow \mathbb{R}$ showing another instance of the same object category. Despite drastic differences of their image pixels, you can recognize both to be related. What renders both images similar although no two pixels are identical? What are the characteristic, salient differences? And how can we obtain a representation $\phi$ that maps images to vectors $\phi(x)$ which retain both, these similarities and also the characteristic differences?
\subsection{Part-based Representation}
    Numerous causes may have led $x$ to be changed into $x'$ (change in articulation, viewpoint, object color or clothing, lighting conditions, 
    etc.). But we can approximate and summarize their effects as a combination of a change in appearance and a change in shape.
    The effect of a change in object shape on an image $x$ can be expressed in terms of a spatial image transformation  $s: \Lambda \rightarrow \Lambda$ acting on the underlying image coordinates, such that the image $x {\circ} s$ depicts the object with altered shape.
    Similarly, we denote the effect of a change in object appearance on an image $x$ as an image transformation $a$ such that the image $a(x)$ depicts the object with altered appearance.

    Note that many of the image changes are local in nature, affecting only part of the image. For instance, animals may only move an individual body part. Similarly, only part of their appearance may vary, \eg, by switching a shirt but not the pants.
    This motivates a part-based factorization of the representation, $\phi(x):=( \phi_1(x), \phi_2(x), \dots )^\top$, so that local changes in appearance and shape stay local and do not alter the overall representation. Nevertheless, global changes can also be accounted for by representing them as as a composition of changes in the individual part representations $\phi_i$.

\subsection{Invariance and Equivariance}
    Let us now carefully observe differences between images $x$ and $x'$ to derive constraints for the representation $\phi$ that is to be learned.
    \emph{i)} Changes in the appearance of an object (\eg in its color or texture), should not impact its shape.
    \emph{ii)} Similarly, changes in shape (\eg through articulation), should not alter the appearance.
    Therefore, the representation needs to separate appearance and shape of the object, so that both can vary individually, i.e. the representation of a part is disentangled into two components $\phi_i(x)=(\alpha_i(x), \sigma_i(x))$.
    Part appearance is modeled as an $n$-dimensional feature vector $\alpha_i(x) \in \mathbb{R}^{n}$.
    Whereas part shape is modeled as a part activation map $\sigma_i(x): \Lambda \rightarrow \mathbb{R}^+$.
    We visualize these maps as colored images (cf. Fig. \ref{fig:architecture}, Fig. \ref{fig:shape}), where each color denotes a single part activation map.

    The invariance of our representation under changes in object appearance and shape can be summarized by the invariance constraints \emph{i)}
    $\alpha_i(x {{\circ}}s) = \alpha_i(x)$ and
    \emph{ii)} $\sigma_i(a(x)) = \sigma_i(x)$.
    In addition, changes in shape should obviously be captured by the shape representation. Thus, for spatial transformations $s$ we obtain the equivariance constraint \emph{iii)}
    $\sigma_i(x {{\circ}}s) = \sigma_i(x) {{\circ}}s$.
    The equivariance constraint simply states that the part activation maps have to consistently track the object part they represent (cf. $\sigma_i(a(x))$ and $\sigma_i(x {\circ} s)$ in Fig. \ref{fig:architecture}).

\subsection{Objective Function for Learning}
    Learning of the representation $\phi$ is driven by integrating invariance and equivariance constraints from the previous section into a reconstruction task.
    The invariance constraints \emph{i}) and \emph{ii}) imply
     \begin{equation}
    \phi_i(x) = [\alpha_i(x), \sigma_i(x)] \overset{!}{=}
    [\alpha_i(x {{\circ}}s), \sigma_i(a({x))]}.
    \label{eq:invarRep}
    \end{equation}
    Let $D([\phi_i(x)]_{i = 1, \dotsc,})$ be a reconstruction of the original image $x$ from the encoded part representations $\phi_1(x), \phi_2(x), ...$ using a decoder $D$. We seek to reconstruct $x$ and, simultaneously, demand the representation to obey the invariance constraints summarized in (\ref{eq:invarRep}),
    \begin{equation}
        \mathcal{L}_\text{rec} =
         \left\lVert x- D\left(\bigl[\alpha_i(x  {{\circ}}s), \sigma_i(a(x))\bigr]_{i = 1,\dotsc}\right)\right\rVert_1
        \label{eq:l_rec}
        .
    \end{equation}
    Moreover, the representation of part shape $\sigma_i(x)$ should be equivariant under deformations. However, simply minimizing equivariance on the scale of pixels, i.e.
    \begin{equation}
     \sum_i \sum_{u \in \Lambda} \Bigl\lVert \sigma_i(x {{\circ}}s)[u]- \sigma_i(x)[s(u)]\Bigr\rVert
     \label{eq:pixeleq}
	    ,
    \end{equation}
    is unstable in practice and favors to the trivial solution of uniform part activations.
    Therefore, we establish an equivariance loss 
    \begin{equation}
    \begin{split}
    \mathcal{L}_{\textrm{equiv}} &= \sum_i \lambda_{\mu} \lVert  \mu[\sigma_i(x{{\circ}}s)]  - \mu[\sigma_i(a(x)) {{\circ}}s]\rVert_{2}  \\ &+ \lambda_{\Sigma} \lVert \Sigma[\sigma_i(x{{\circ}}s)] -  \Sigma[\sigma_i(a(x)) {{\circ}}s] \rVert_{1} \;,
    \end{split}
    \label{eq:equiv}
    \end{equation}
    where $\mu[\sigma_i(x)]$ and $\Sigma[\sigma_i(x)]$ denote the mean and covariance over coordinates of
    $\sigma_i(x)/ \sum_{u \in \Lambda} \sigma_i(x)[u]$.
    Note that we have employed invariance \emph{ii)} so that we can use the same shape encoding $\sigma_i(a(x))$ as in (\ref{eq:l_rec}).
    The overall training objective of our model is to minimize the reconstruction and equivariance loss,
    \begin{equation}
        \mathcal{L} =  \mathcal{L}_\text{rec} + \mathcal{L}_\text{equiv}
        \label{eq:totalloss}
        .
    \end{equation}

    Note that object parts a priori unknown, but in order to reconstruct the object, the representations $\phi_i$ automatically learn to structure it into meaningful parts which capture the variance in shape and appearance. In particular, we do not need to introduce artificial prior assumptions about the relations between parts, such as the separation constraints employed in \cite{Zhang:2018vz, Thewlis:2017wi}.
    Instead, the local modelling of the part representation (cf. sec. \ref{sec:architecture}) as disentangled components of shape and appearance drives our representation to meaningfully structure the object and learn natural relations between parts.

\subsection{Unsupervised Learning of Part-based Shape and Appearance}
    \label{sec:architecture}
    Subsequently, we discuss the network architecture in Fig. \ref{fig:architecture} for unsupervised learning of an appearance and shape representation using the reconstruction (\ref{eq:l_rec}) and equivariance loss (\ref{eq:equiv}).
    Learning considers image pairs $x {{\circ}} s$ and  $a(x)$.
    The leading design principle of our architecture is to model the local interplay between part shape and part appearance.
    In a fully differentiable procedure equivariance of part activation maps is used to extract part appearances from $x {{\circ}} s$ and assign them to corresponding image regions in $x$.

    \textbf{Part Shape.}
    In a shape stream (cf. Fig. \ref{fig:architecture}), an hourglass network \cite{Newell:2016vq} $E_{\sigma}$ learns to \emph{localize} parts $i$ by means of part activation maps $\sigma_i(a(x)) \in \mathbb{R}^{h\times w}$.
    The hourglass model nicely suits this task, since it preserves pixel-wise locality and integrates information from multiple scales \cite{Newell:2016vq}. Multi-scale context is essential to learn the relations between parts and consistently assign them to an object.

    \textbf{Part Appearance.}
    Let us now localize the parts by detecting $\sigma_i(x {\circ} s)$ in a spatially transformed image $x {\circ} s$ using the same network $E_{\sigma}$ (cf. Fig. \ref{fig:architecture} Appearance Stream).
    To learn representations of part appearance $\alpha_i(x {\circ} s)$, we first stack all normalized part activations $\sigma_i(x {\circ} s)/\sum_{u \in \Lambda} \sigma_i(x {\circ} s)[u]$ and an image encoding, i.e., the output of the first convolution filters of the network $E_\sigma$ applied to $x{\circ} s$. A second hourglass network $E_{\alpha}$ takes this stack as input and maps it onto a localized image appearance encoding ${f_{x{\circ} s}} \in \mathbb{R}^{h\times w \times n}$.
    To obtain local part appearances, we average pool these features at all locations where part $i$ has positive activation
    \begin{equation}
    \alpha_i(x {\circ} s) =
    \frac{ \sum_{u \in  \Lambda} f_{x{\circ} s}[u] \sigma_i(x {\circ} s)[u] }
    { \sum_{u \in  \Lambda} \sigma_i(x {\circ} s)[u] }
    \label{eq:appearance_encoding}.
    \end{equation}

    \textbf{Reconstructing the Original Image.}
    Next we reconstruct $x$ from part appearances 
    $\alpha_i(x {\circ} s)$ and 
    part activations $\sigma_i(a(x))$ using a U-Net~\cite{Ronneberger:2015gk} (cf. Fig. \ref{fig:architecture}). The encoder of the U-Net is simply a set of fixed downsampling layers. Only its decoder is learned.
    We approximate part activations 
    $\sigma_i(a(x))$ by their
    first two moments,
    \begin{equation}
    \tilde{\sigma}_i(a(x))[u] = \frac{1}{1 + (u -\mu_i)^T \Sigma_i^{-1} (u - \mu_i)},
    \end{equation}
    where $\mu_i$ and  $\Sigma_i$ denote the mean and covariance of the normalized part activation maps $\sigma_i(a(x))/\sum_{u \in \Lambda} \sigma_i(a(x))[u]$. Thus, extra information present in part activations 
    is neglected, forcing the shape encoder $E_\sigma$ to concentrate on an unambiguous part localization (or else reconstruction loss would increase).
    The second input to the decoder $D$ in Eq.~\ref{eq:l_rec} are part appearances $\alpha_i(x{\circ} s)$.
    Note that $\alpha_i(x{\circ} s)$ are feature vectors devoid of localization.
    We exploit the fact that the corresponding part activations 
     $\tilde{\sigma}_i(a(x))$ designate the regions of parts $i$ in image $x$ (cf. Fig \ref{fig:architecture}) to project the part appearances onto a localized appearance encoding $f_x$:
    \begin{equation}
        f_x[u]=  \sum_i
        \frac{ \alpha_i(x{\circ} s) \cdot \tilde{\sigma}_i(a(x))[u]}
        {1+ \sum_j \tilde{\sigma}_j(a(x))[u]}
        .
    \end{equation}
    To reconstruct $x$, the U-Net can then exploit the local correspondence between $f_x$,  $\tilde{\sigma}_i(a(x))$ and $x$.

\subsection{Implementation Details}
For appearance transformation $a$ we apply changes in brightness, contrast, and hue.
For image datasets, $s$ are thin plate spline (TPS) transformations.
On video datasets, in addition to applying synthetic TPS transformations we randomly sample another frame from the same video sequence which acts as $x {\circ} s$.
Selecting the number of parts is uncritical, since our model is robust for different numbers of parts, Tab. \ref{tab:static}.
For image synthesis in Sect. \ref{sec:dis} we train the decoder $D$ with an adversarial loss \cite{Isola2017image}.
We refer to the supplementary for further details on the architecture and the experimental setup.

\section{Experiments}
	In this section we evaluate our unsupervised approach for learning disentangled representation of appearance and shape.
	Sect. \ref{sec:shape} evaluates and visualizes the shape representation on the task of unsupervised landmark discovery.
	Sect. \ref{sec:dis} investigates the disentangling of our representation.
	On the task of conditional image generation, we compare our unsupervised shape/appearance disentanglement performance against a state-of-the-art disentangling method that utilizes groundtruth shape annotations.
	Moreover, on the task of frame-to-frame video translation we show the robustness of our representation across multiple frames.
	Additionally, we evaluate the ability of our method to disentangle parts and their local appearance and shape using a part-wise appearance transfer.

\subsection{Datasets}
    \begin{table}
	\caption{Difficulties of datasets: articulation, intra-class variance, background clutter and viewpoint variation.}
	\label{tab:challenges}
	\begin{tabular}{l|rrrr}
		Dataset &  Articul.& Var. &  Backgr.& Viewp.  \\ \hline
		CelebA &   &  &  &    \\
		Cat Head & &  \checkmark&  &   \\
		CUB-200-2011 & & \checkmark& \checkmark&   \\
		Human3.6M &\checkmark& &  & \checkmark  \\
		BBC Pose &  \checkmark&  & \checkmark&  \\
		Dogs Run & \checkmark& \checkmark& \checkmark&   \\
		Penn Action & \checkmark& \checkmark& \checkmark& \checkmark\\
	\end{tabular}
\end{table}

\begin{figure}[t]
	\begin{subfigure}{0.5\textwidth}
	\centering
	\includegraphics[trim={0cm 0cm 0cm 0cm},clip, width=1.\linewidth]{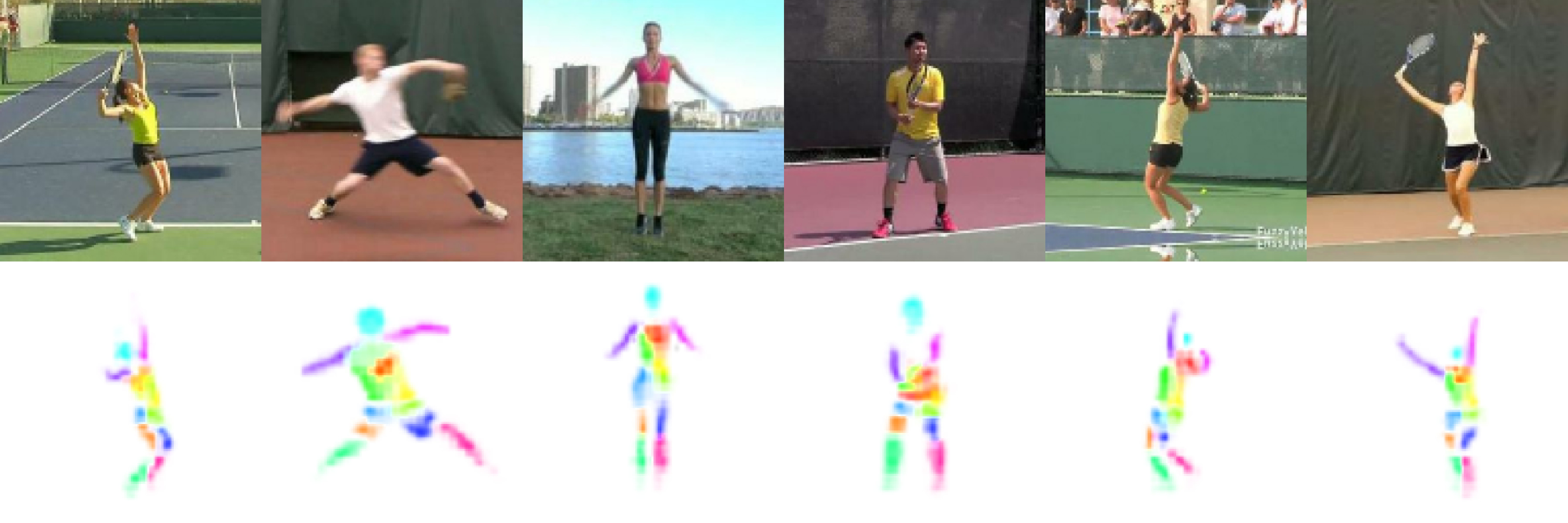}\caption{}
	\label{fig:shape_penn}
	\end{subfigure}
	\begin{subfigure}{0.5\textwidth}
	\centering
	\includegraphics[trim={0cm 0cm 0cm 0cm},clip, width=1.\linewidth]{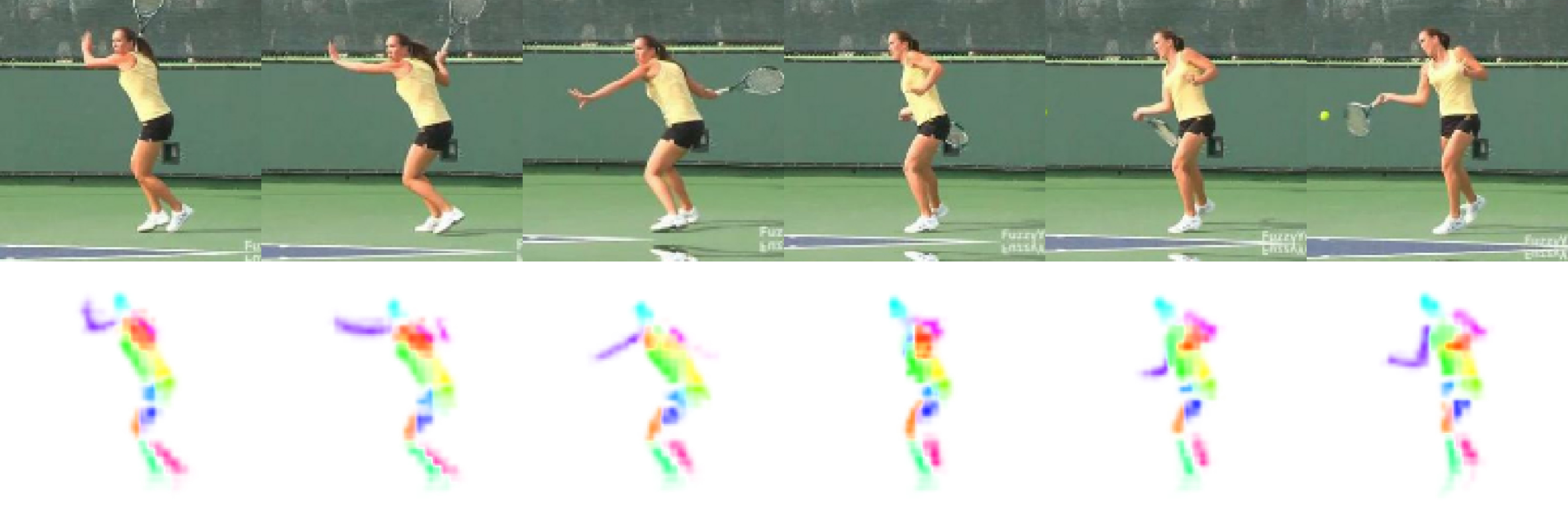}\caption{}
	\label{fig:shape_tennis}
	\end{subfigure}
	\caption{Learned shape representation on Penn Action. For visualization, 12 of 16 part activation maps are plotted in one image. (a) Different instances, showing intra-class consistency and (b) video sequence, showing consistency and smoothness under motion, although each frame is processed individually.}
	\label{fig:shape}
\end{figure}

	\textbf{CelebA} \cite{Liu:2015vj} contains ca. 200k celebrity faces of 10k identities.
	We resize all images to $128\times 128$ and exclude the training and test set of the MAFL subset, following \cite{Thewlis:2017wi}.
	As  \cite{Thewlis:2017wi, Zhang:2018vz}, we train the regression (to 5 ground truth landmarks) on the MAFL training set (19k images) and test on the MAFL test set (1k images).

	\textbf{Cat Head} \cite{Zhang:2008uj}  has nearly 9k images of cat heads.
	We use the train-test split of \cite{Zhang:2018vz} for training (7,747 images) and testing (1,257 images).
	We regress 5 of the 7 (same as \cite{Zhang:2018vz}) annotated landmarks.
	The images are cropped by bounding boxes constructed around the mean of the ground truth landmark coordinates and resized to $128\times128$.

	\textbf{CUB-200-2011} \cite{Wah:2011vq} comprises ca. 12k images of birds in the wild from 200 bird species.
	We excluded bird species of seabirds, roughly cropped using the provided landmarks as bounding box information and resized to $128\times128$.
	We aligned the parity with the information about the visibility of the eye landmark.
	For comparing with \cite{Zhang:2018vz} we used their published code.

	\textbf{BBC Pose} \cite{Charles:2013tb} contains videos of sign-language signers with varied appearance in front of a changing background. Like \cite{Jakab:2018wc} we loosely crop around the signers.
	The test set includes 1000 frames and the test set signers did not appear in the train set.
	For evaluation, as \cite{Jakab:2018wc}, we utilized the provided evaluation script, which measures the PCK around $d=6$ pixels in the original image resolution.

	\textbf{Human3.6M} \cite{Ionescu:2014ua} features human activity videos.
	We adopt the training and evaluation procedure of \cite{Zhang:2018vz}.
	For proper comparison to \cite{Zhang:2018vz} we also removed the background using the off-the-shelf unsupervised background subtraction method provided in the dataset.

	\textbf{Penn Action} \cite{Zhang:2013tr} contains 2326 video sequences of 15 different sports categories.
	For this experiment we use 6 categories (tennis serve, tennis forehand, baseball pitch, baseball swing, jumping jacks, golf swing).
	We roughly cropped the images around the person, using the provided bounding boxes, then resized to $128\times128$.

	\textbf{Dogs Run} is made from dog videos from YouTube totaling in 1250 images under similar conditions as in Penn Action. The dogs are running in one direction in front of varying backgrounds. The 17 different dog breeds exhibit widely varying appearances.

	\textbf{Deep Fashion} \cite{Liu:2016vv} consists of ca. 53k in-shop clothes images in high-resolution of $256 \times 256$. We selected the images which are showing a full body (all keypoints visible, measured by \cite{Cao:2017vv}) and used the provided train-test split.
	For comparison with Esser \etal \cite{Esser:2018ue} we used their published code.

	\subsection{Evaluating Unsupervised Learning of Shape}\label{sec:shape}
	\begin{figure}[t]
	\centering
	\includegraphics[trim={0cm 0cm 0cm 0cm},clip, width=1.\linewidth]{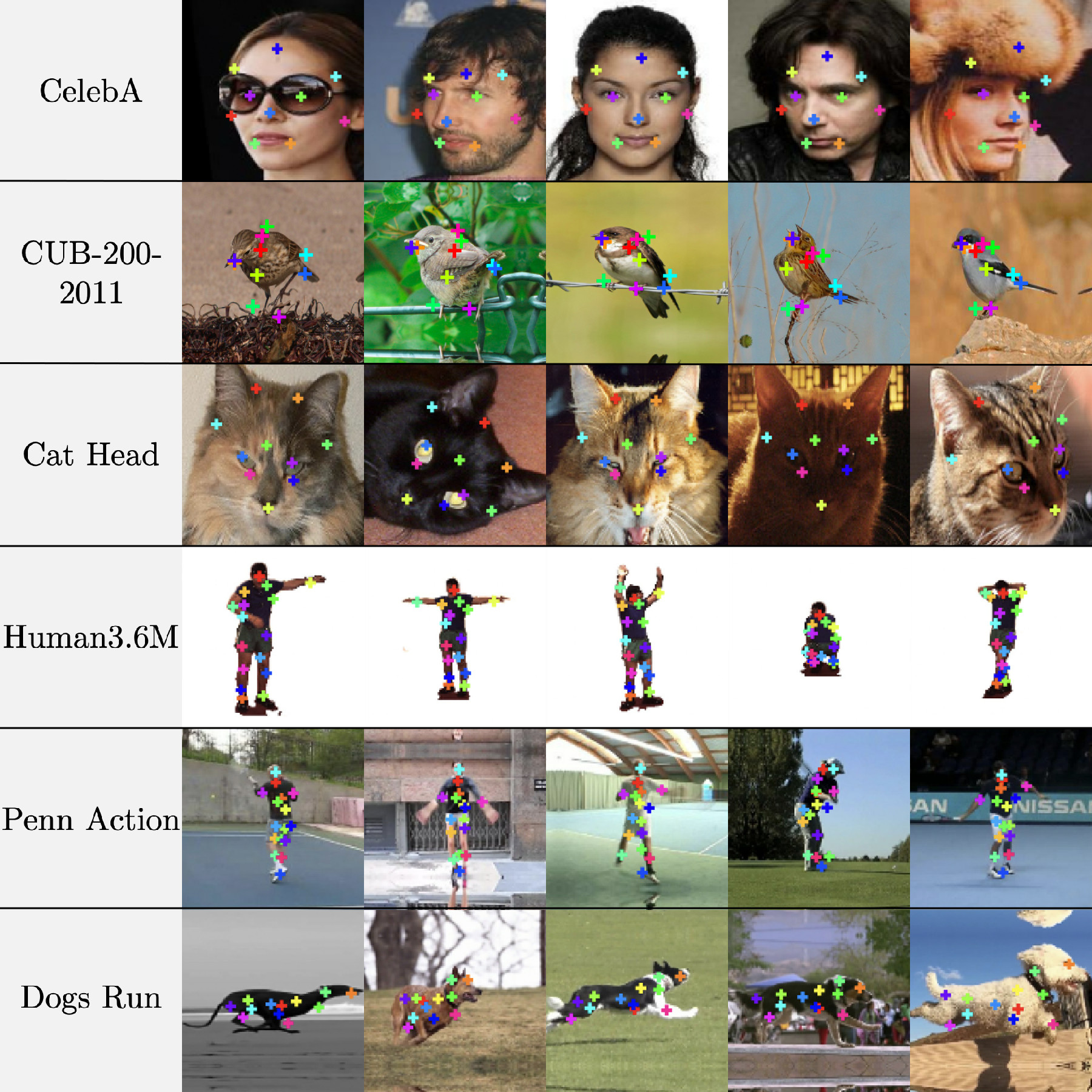}
	\caption{{Unsupervised discovery of landmarks on diverse object classes such as human or cat faces and birds and for highly articulated human bodies and running dogs.}}
	\label{fig:kp_overview}
\end{figure}

\begin{figure}[t]
	\centering
	\includegraphics[trim={0cm 0cm 0cm 0cm},clip, width=1.\linewidth]{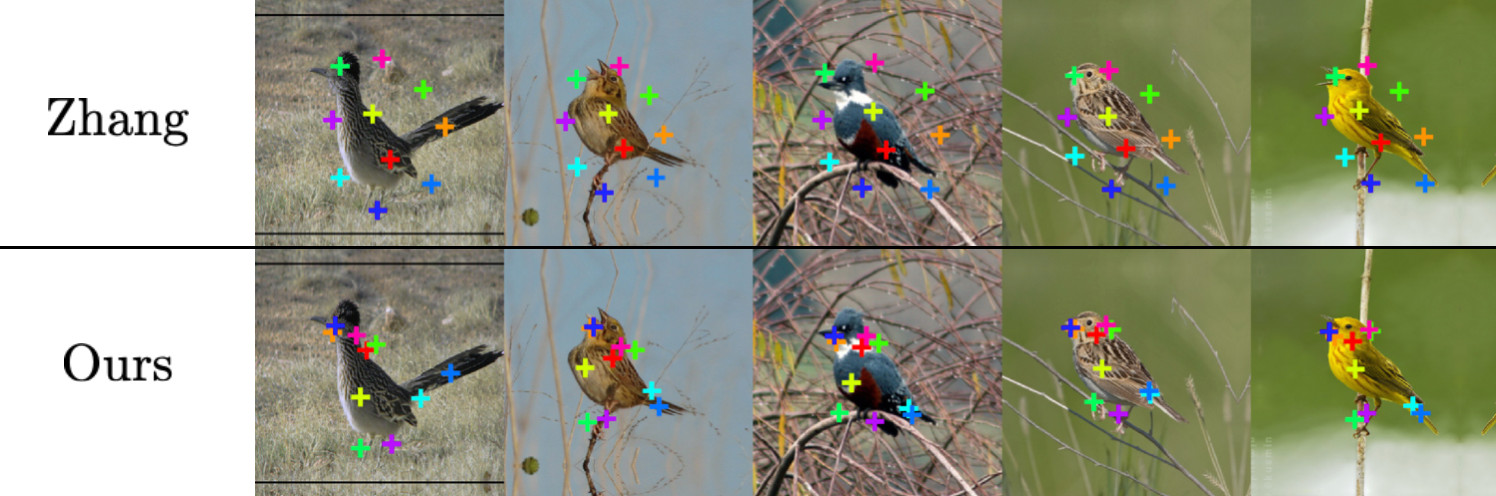}
	\caption{Comparing discovered keypoints against \cite{Zhang:2018vz} on CUB-200-2011. We improve on object coverage and landmark consistency. Note our flexible part placement compared to a rather rigid placement of \cite{Zhang:2018vz} due to their part separation bias.}
	\label{fig:compare}
\end{figure}

\begin{table}[t]
	\caption{{Error of unsupervised methods for landmark prediction on the Cat Head, MAFL (subset of CelebA), and CUB-200-2011 testing sets.
	The	error is in \% of inter-ocular distance for Cat Head and MAFL and in \% of edge length of the image for CUB-200-2011.}}
	\label{tab:static}
	\centering
	\begin{tabular}{l|cccc}
	\hline
	Dataset & Cat Head &  & MAFL & CUB\\
	  \# Landmarks &10 & 20  & 10 &10  \\
	  \hline
	 Thewlis \cite{Thewlis:2017wi}
	 & 26.76 & 26.94 & 6.32  & -  \\
	 Jakab \cite{Jakab:2018wc}
	 & - & - & \textbf{3.19} & - \\
	 Zhang \cite{Zhang:2018vz}
	 & 15.35 & 14.84 & 3.46 & 5.36 \\
	  Ours & \textbf{9.88}  & \textbf{9.30} & 3.24 & \textbf{3.91}  \\ \hline  
	\end{tabular}
\end{table}

	Fig. \ref{fig:shape} visualizes the learned shape representation.
	To quantitatively evaluate the shape estimation, we measure how well groundtruth landmarks (only during testing) are predicted from it.
	The part means $\mu[\sigma_i(x)]$ (cf. (\ref{eq:equiv})) serve as our landmark estimates and we measure the error when linearly regressing the human-annotated groundtruth landmarks from our estimates.
	For this, we follow the protocol of Thewlis \etal \cite{Thewlis:2017wi}, fixing the network weights after training the model, extracting unsupervised landmarks and training a single linear layer without bias.
	The performance is quantified on a test set by the mean error and the percentage of correct landmarks (PCK).
	We extensively evaluate our model on a diverse set of datasets, each with specific challenges. An overview over the challenges implied by each dataset is given in Tab. \ref{tab:challenges}.
	On all datasets except for MAFL we outperform the state-of-the-art by a significant margin.

	\textbf{Diverse Object Classes.}
	On the object classes of human faces, cat faces, and birds (datasets CelebA, Cat Head, and CUB-200-2011) our model predicts landmarks consistently across different instances, cf. Fig. \ref{fig:kp_overview}.
	Tab. \ref{tab:static} compares against the state-of-the-art. Due to different breeds and species the Cat Head, CUB-200-2011 exhibit large variations between instances. Especially on these challenging datasets we outperform competing methods by a large margin.
	Fig. \ref{fig:compare} also provides a direct visual comparison to \cite{Zhang:2018vz} on CUB-200-2011. It becomes evident that our predicted landmarks track the object much more closely. In contrast, \cite{Zhang:2018vz} have learned a slightly deformable, but still rather rigid grid.
	This is due to their separation constraint, which forces landmarks to be mutually distant. We do not need such a problematic bias in our approach, since the localized, part-based representation and reconstruction guides the shape learning and captures the object and its articulations more closely.
	% BBC POSE Results
\begin{table}[t]
	\caption{{
	Performance of landmark prediction on BBC Pose test set. As upper bound, we also report the performance of supervised methods.
	The metric is \% of points within 6 pixels of groundtruth location. 
	}}
	\label{tab:bbcpose}
	\centering
	\begin{tabular}{ll|cr}
	\hline
	BBC Pose &   &    { Accuracy}  \\
	 \hline
	supervised & Charles \cite{Charles:2013tb} &
	   79.9\%  \\ % 79.90
	 & Pfister \cite{Pfister:2015uo}  &
	  88.0\%  \\ \hline % 88.01
	unsupervised &Jakab \cite{Jakab:2018wc} &
	 68.4\%  \\  % 68.44
	  &Ours &  \textbf{74.5}\% \\

	\hline
	\end{tabular}
\end{table}
%
% HUMAN3.6M Results
\begin{table}[t]
	\caption{{Comparing against supervised, semi-supervised and unsupervised methods for landmark prediction on the Human3.6M test set. The
	error is in \% of the edge length of the image. All methods predict 16 landmarks.
	}}
	\label{tab:human}
	\centering
	\begin{tabular}{ll|cr}
	\hline
	 Human3.6M   & &  { Error w.r.t. image size}  \\
	 \hline
	 supervised & Newell \cite{Newell:2016vq}
	  &2.16  \\  \hline
	 semi-supervised & Zhang \cite{Zhang:2018vz}
	  & 4.14  \\ \hline
	 unsupervised & Thewlis \cite{Thewlis:2017wi}
	 & 7.51  \\
	  & Zhang \cite{Zhang:2018vz}
		& 4.91 \\
	  & Ours& \textbf{2.79} \\
	\hline
	\end{tabular}
\end{table}

	\textbf{Articulated Object Pose.}
	Object articulation makes consistent landmark discovery challenging.
	Fig. \ref{fig:kp_overview} shows that our model exhibits strong landmark consistency under articulation and covers the full human body meaningfully.
	Even fine-grained parts such as the arms are tracked across heavy body articulations, which are frequent in the Human3.6M and Penn Action datasets.
    Despite further complications such as viewpoint variations or blurred limbs our model can detect landmarks on Penn Action.
	Additionally, complex background clutter as in BBC Pose and Penn Action, does not hinder finding the object.
	Experiments on the Dogs Run dataset underlines that even completely dissimilar dog breeds can be related via semantic parts.
	Tab. \ref{tab:bbcpose} and Tab. \ref{tab:human} summarize the quantitative evaluations: we outperform other unsupervised and semi-supervised methods by a large margin on both datasets.
	On Human3.6M, our approach achieves a large performance gain even compared to methods that utilize optical flow supervision.
	On BBC Pose, we outperform \cite{Jakab:2018wc} by $6.1\%$, reducing the performance gap to supervised methods significantly.
\subsection{Disentangling Shape and Appearance}\label{sec:dis}
	\begin{figure}[t]
	\centering
	\includegraphics[trim={0cm 0cm 0cm 0cm},clip, width=1.\linewidth]{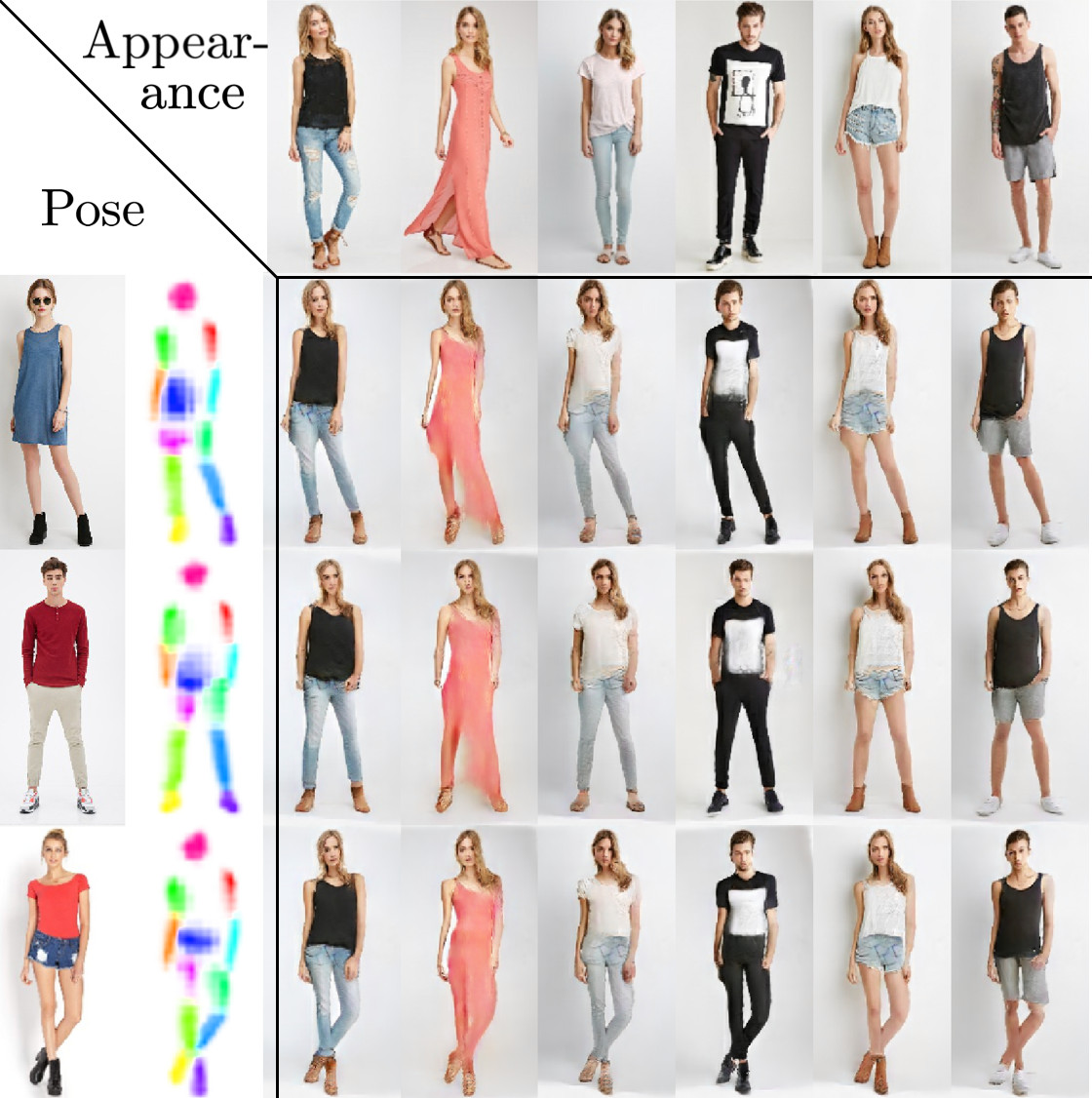}
	\caption{Transferring shape and appearance on Deep Fashion. Without annotation the model estimates shape (12 out of 16 part activations shown), 2nd column. Target appearance is extracted from images in top row to synthesize images. Note that we trained without image pairs only using synthetic transformations.
	All images are from the test set.}
	\label{fig:allswaps}
\end{figure}

	Disentangled representations of object shape and appearance allow to alter both properties individually to synthesize new images. The ability to flexibly control the generator allows, for instance, to change the pose of a person or their clothing. In contrast to previous work \cite{Esser:2018ue, Denton:2017uf, Ma:2017uu, Ma:2017wq, deBem:2018wp, Jakab:2018wc},
	we achieve this ability without requiring supervision \textit{and} using a flexible part-based model instead of a holistic representation. This allows to explicitly control the parts of an object that are to be altered. We quantitatively compare against \emph{supervised} state-of-the-art disentangled synthesis of human figures. Also we qualitatively evaluate our model on unsupervised synthesis of still images, video-to-video translation, and local editing for appearance transfer.

	\textbf{Conditional Image Generation.}
	On Deep Fashion \cite{Liu:2015vj, Liu:2016vv}, a benchmark dataset for supervised disentangling methods, the task is to separate person ID (appearance) from body pose (shape) and then synthesize new images for previously unseen persons from the test set in eight different poses. We randomly sample the target pose and appearance conditioning from the test set. Fig. \ref{fig:allswaps} shows qualitative results.
	We quantitatively compare against supervised state-of-the-art disentangling \cite{Esser:2018ue} by evaluating \emph{i)} invariance of appearance against variation in shape by the re-identification error and \emph{ii)} invariance of shape against variation in appearance by the distance in pose between generated and pose target image.
	\begin{table}
	\caption{Mean average precision (mAP) and rank-n accuracy for person re-identification on synthesized images after performing shape/appearance swap. Input images from Deep Fashion test set. Note \cite{Esser:2018ue} is supervised w.r.t. shape.}
	\label{tab:reid}
	\begin{tabular}{l|cccr}
		\hline
		& mAP & rank-1 & rank-5 & rank-10 \\ \hline
		VU-Net \cite{Esser:2018ue} & 88.7\% & 87.5\% & {98.7}\% & {99.5}\% \\
		Ours & {90.3}\% & {89.4}\% &{98.2}\% & {99.2}\% \\ \hline
	\end{tabular}
\end{table}
\begin{table}
	\caption{Percentage of Correct Keypoints (PCK) for pose estimation on shape/appearance swapped generations.\;$\alpha$ is pixel distance divided by image diagonal. Note that \cite{Esser:2018ue} serves as upper bound, as it uses the groundtruth shape estimates.}
	\label{tab:pose}
	\begin{tabular}{l|cccr}
		\hline
		$\alpha$ & $2.5\%$ &  $5\%$ & $7.5\%$ & $10\%$ \\ \hline
		VU-Net \cite{Esser:2018ue} & {95.2}\% & {98.4}\% & {98.9}\% & {99.1}\% \\
		Ours & 85.6\% & 94.2\% &96.5\% & 97.4\% \\ \hline
	\end{tabular}
\end{table}

\begin{figure}[t]
	\centering
	\includegraphics[trim={0cm 0cm 0cm 0cm},clip, width=1.\linewidth]{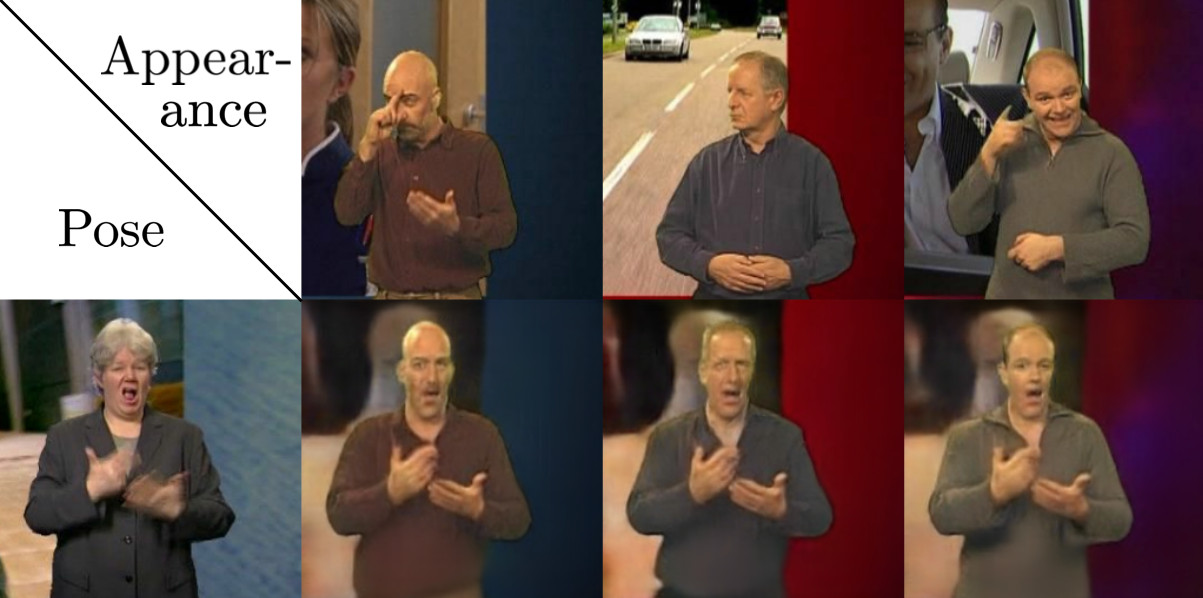}
	\caption{Video-to-video translation on BBC Pose. Top-row: target appearances, left: target pose.
	Note that even fine details in shape are accurately captured. Visit the project page for the video.}
	\label{fig:bbcthumb}
\end{figure}

	\emph{i)} To evaluate appearance we fine-tune an ImageNet-pretrained \cite{Russakovsky2015imagenet} Inception-Net \cite{Szegedy2015inception} with a re-identification (ReID) algorithm \cite{Xiao:2017} via a triplet loss \cite{Hermans:2017} to the Deep Fashion training set.
	On the generated images we evaluate the standard metrics for ReID, mean average precision (mAP) and rank-1, -5, and -10 accuracy in Tab. \ref{tab:reid}.
	Although our approach is unsupervised it is competitive compared to the supervised VU-Net \cite{Esser:2018ue}.
	\\
	\emph{ii)} To evaluate shape, we extract keypoints using the pose estimator \cite{Cao:2017vv}. Tab. \ref{tab:pose} reports the difference between generated and pose target in percentage of correct keypoints (PCK). As would be expected, VU-Net performs better, since it is trained with exactly the keypoints of \cite{Cao:2017vv}. Still our approach achieves an impressive PCK without supervision underlining the disentanglement of appearance and shape.

	\textbf{Video-to-Video Translation.}
	To evaluate the robustness of our disentangled representation, we synthesize a video sequence frame-by-frame without temporal consistency constraints. On BBC Pose \cite{Charles:2013tb}, one video provides a sequence of target poses, another video a sequence of source appearances to then perform retargeting, Fig. \ref{fig:bbcthumb}.
	Although there is no temporal coupling, the generated sequences are smooth and pose estimation is robust. Secondly, the training on the natural spatial deformations in video data enables the model to encapsulate realistic transitions such as out-of-plane rotation and complex 3D articulation of hands and even fingers.
	Due to the local nature of the part based representation, the model is robust to variations in the background and focuses on the object whilst the background is only roughly reconstructed.
	
	\textbf{Part Appearance Transfer.}
	The flexible part-based representation allows to explicitly control local appearance. Fig. \ref{fig:partswaps} shows swaps of appearance for shirt, pants, etc. In contrast to holistic representations \cite{Esser:2018ue, Jakab:2018wc, Ma:2017uu, Ma:2017wq, deBem:2018wp}, we can guarantee the transfer to be focused on selected object parts.
	\begin{figure}[t]
	\begin{subfigure}{0.49\linewidth}
	\centering
	\includegraphics[trim={0cm 0cm 0cm 0cm},clip, width=1.\linewidth]{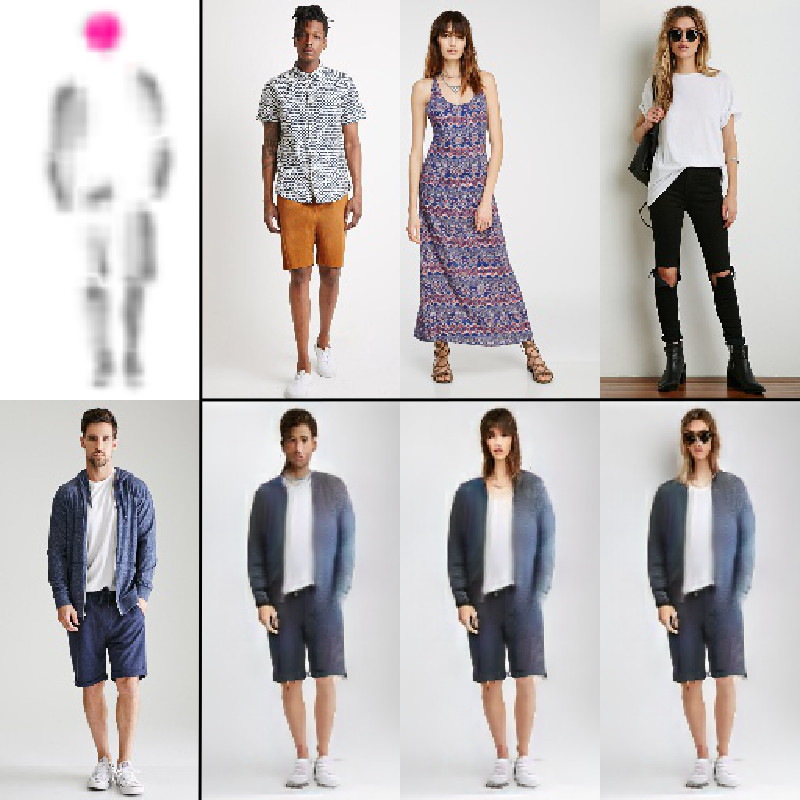}\caption{}
	\label{fig:part3_00}
	\end{subfigure}
	\begin{subfigure}{0.49\linewidth}
	\centering
	\includegraphics[trim={0cm 0cm 0cm 0cm},clip, width=1.\linewidth]{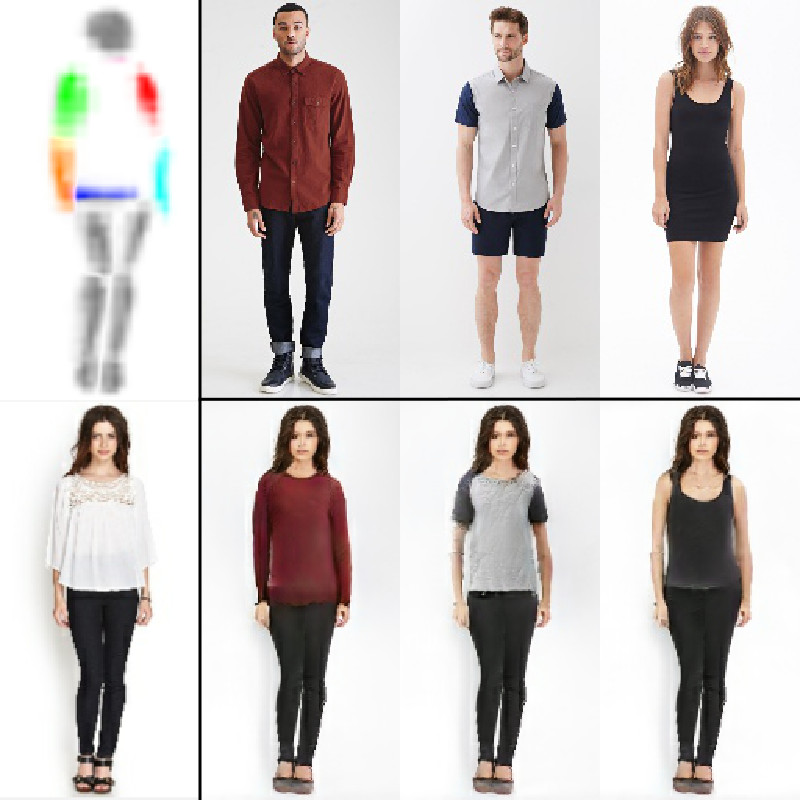}\caption{}
	\label{fig:part3_11}
	\end{subfigure}
	\begin{subfigure}{0.49\linewidth}
	\centering
	\includegraphics[trim={0cm 0cm 0cm 0cm},clip, width=1.\linewidth]{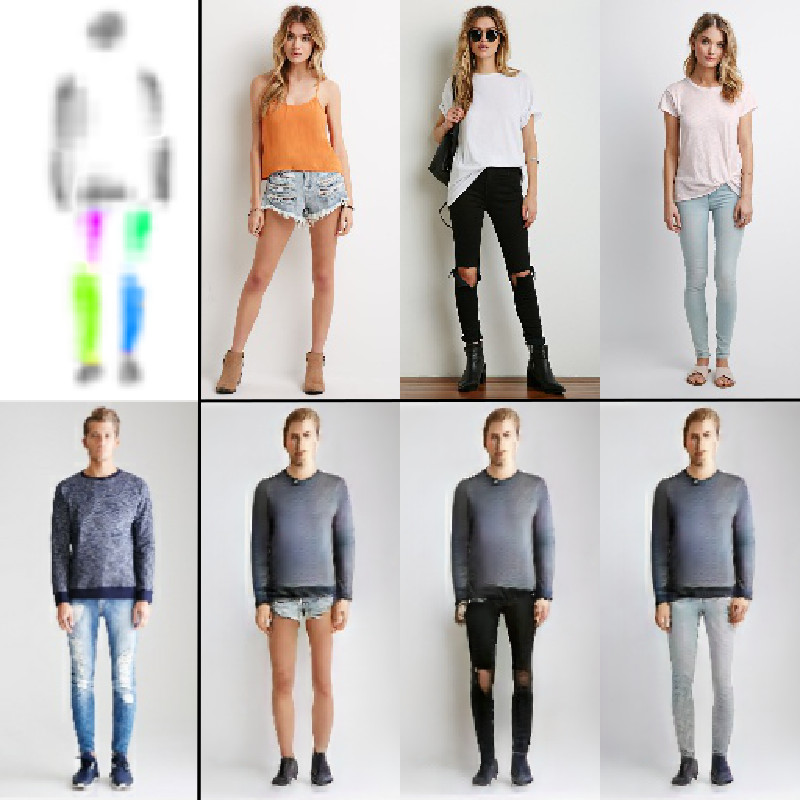}\caption{}
	\label{fig:part3_21}
	\end{subfigure}
	\begin{subfigure}{0.49\linewidth}
	\centering
	\includegraphics[trim={0cm 0cm 0cm 0cm},clip, width=1.\linewidth]{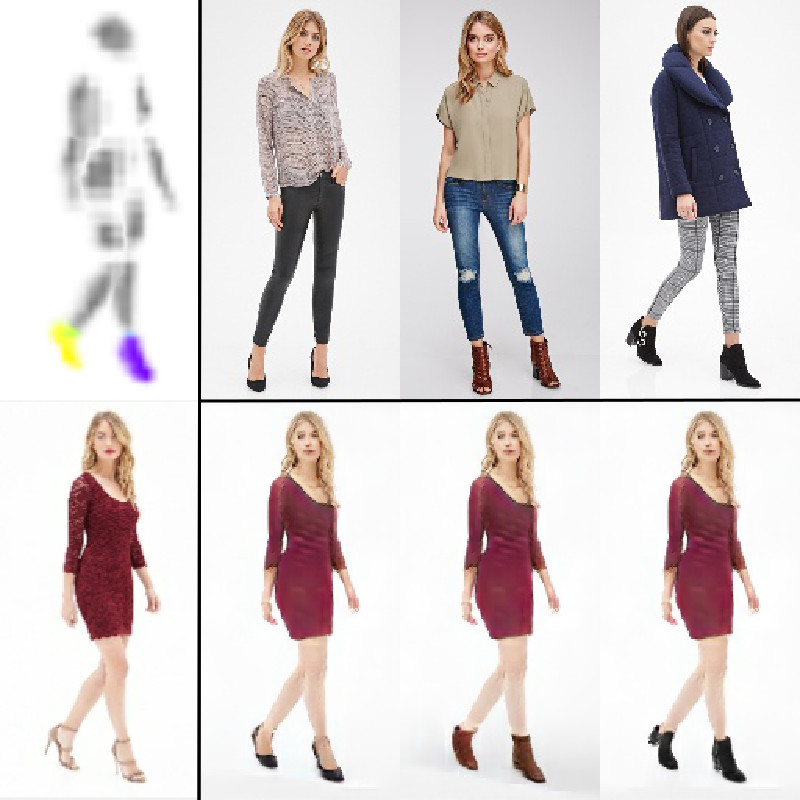}\caption{}
	\label{fig:part3_30}
	\end{subfigure}
	\caption{Swapping part appearance on Deep Fashion. Appearances can be exchanged for parts individually and without altering shape. We show part-wise swaps for (a) head (b) torso (c) legs, (d) shoes. All images are from the test set. 12 out of 16 part shapes are visualized in the top left image.}
	\label{fig:partswaps}
\end{figure}

\section{Conclusion}
We have presented an unsupervised approach to learning the compositional part structure of objects by disentangling shape from appearance. We incorporate invariance and equivariance constraints in a generative framework. The model discovers consistent parts without requiring prior assumptions. Experiments show our approach significantly improves upon previous unsupervised methods.
\blfootnote{This work has been supported in part by DFG grant OM81/1-1
and a hardware donation from NVIDIA Corporation.}
\newpage
{\small
\bibliographystyle{ieee}
\bibliography{mend2}

\begin{thebibliography}{10}\itemsep=-1pt

\bibitem{Balakrishnan:2018wo}
G.~Balakrishnan, A.~Zhao, A.~V. Dalca, F.~Durand, and J.~V. Guttag.
\newblock {Synthesizing images of humans in unseen poses}.
\newblock {\em arXiv preprint arXiv:1804.07739}, 2018.

\bibitem{Walidtf}
W.~Benbihi.
\newblock Stacked hourglass model.
\newblock \url{https://github.com/wbenbihi/hourglasstensorlfow}, 2017.

\bibitem{Bengio2013rep}
Y.~Bengio, A.~Courville, and P.~Vincent.
\newblock {Representation learning: A review and new perspectives}.
\newblock {\em TPAMI}, 2013.

\bibitem{Cao:2017vv}
Z.~Cao, T.~Simon, S.-E. Wei, and Y.~Sheikh.
\newblock {Realtime multi-person 2d pose estimation using part affinity
  fields}.
\newblock In {\em CVPR}, 2017.

\bibitem{Charles:2013tb}
J.~Charles, T.~Pfister, D.~R. Magee, D.~C. Hogg, and A.~Zisserman.
\newblock {Domain adaptation for upper body pose tracking in signed tv
  broadcasts}.
\newblock In {\em BMVC}, 2013.

\bibitem{Chen2016infogan}
X.~Chen, Y.~Duan, R.~Houthooft, J.~Schulman, I.~Sutskever, and P.~Abbeel.
\newblock {Infogan: Interpretable representation learning by information
  maximizing generative adversarial nets}.
\newblock {\em NIPS}, 2016.

\bibitem{Cootes:1998tn}
T.~F. Cootes, G.~J. Edwards, and C.~J. Taylor.
\newblock {Active appearance models}.
\newblock In {\em ECCV}, 1998.

\bibitem{deBem:2018wp}
R.~de~Bem, A.~Ghosh, T.~Ajanthan, O.~Miksik, N.~Siddharth, and P.~H.~S. Torr.
\newblock {Dgpose: Disentangled semi-supervised deep generative models for
  human body analysis}.
\newblock {\em arXiv preprint arXiv:1804.06364}, 2018.

\bibitem{Denton:2017uf}
E.~L. Denton and V.~Birodkar.
\newblock {Unsupervised learning of disentangled representations from video}.
\newblock In {\em NIPS}, 2017.

\bibitem{Desjardins2012dr}
G.~Desjardins, A.~Courville, and Y.~Bengio.
\newblock Disentangling factors of variation via generative entangling.
\newblock {\em arXiv preprint arXiv:1210.5474}, 2012.

\bibitem{Eastwood2018dr}
C.~Eastwood and C.~K. Williams.
\newblock A framework for the quantitative evaluation of disentangled
  representations.
\newblock {\em ICLR}, 2018.

\bibitem{eigenstetter:CVPR:2014}
A.~Eigenstetter, M.~Takami, and B.~Ommer.
\newblock Randomized max-margin compositions for visual recognition.
\newblock In {\em Proceedings of the IEEE Conference on Computer Vision and
  Pattern Recognition}, pages 3590--3597. IEEE, IEEE, 2014.

\bibitem{Esser:2018ue}
P.~Esser, E.~Sutter, and B.~Ommer.
\newblock {A variational u-net for conditional appearance and shape
  generation}.
\newblock {\em CVPR}, 2018.

\bibitem{Felzenszwalb:2010ve}
P.~F. Felzenszwalb, R.~B. Girshick, D.~A. McAllester, and D.~Ramanan.
\newblock {Object detection with discriminatively trained part-based models}.
\newblock {\em TPAMI}, 2010.

\bibitem{Fergus2003object}
R.~Fergus, P.~Perona, and A.~Zisserman.
\newblock Object class recognition by unsupervised scale-invariant learning.
\newblock In {\em CVPR}, 2003.

\bibitem{Fischler1973rep}
M.~A. Fischler and R.~A. Elschlager.
\newblock The representation and matching of pictorial structures.
\newblock {\em IEEE Transactions on computers}, 1973.

\bibitem{Hermans:2017}
A.~Hermans, L.~Beyer, and B.~Leibe.
\newblock In defense of the triplet loss for person re-identification.
\newblock {\em arXiv preprint arXiv:1703.07737}, 2017.

\bibitem{Higgins2016betavae}
I.~Higgins, L.~Matthey, A.~Pal, C.~Burgess, X.~Glorot, M.~Botvinick,
  S.~Mohamed, and A.~Lerchner.
\newblock Beta-vae: Learning basic visual concepts with a constrained
  variational framework.
\newblock {\em ICLR}, 2017.

\bibitem{Ionescu:2011ue}
C.~Ionescu, F.~Li, and C.~Sminchisescu.
\newblock {Latent structured models for human pose estimation}.
\newblock In {\em ICCV}, 2011.

\bibitem{Ionescu:2014ua}
C.~Ionescu, D.~Papava, V.~Olaru, and C.~Sminchisescu.
\newblock {Human3.6m: Large scale datasets and predictive methods for 3d human
  sensing in natural environments}.
\newblock {\em TPAMI}, 2014.

\bibitem{Isola2017image}
P.~Isola, J.-Y. Zhu, T.~Zhou, and A.~A. Efros.
\newblock Image-to-image translation with conditional adversarial networks.
\newblock {\em arXiv preprint}, 2017.

\bibitem{Jakab:2018wc}
T.~Jakab, A.~Gupta, H.~Bilen, and A.~Vedaldi.
\newblock {Conditional image generation for learning the structure of visual
  objects}.
\newblock {\em NIPS}, 2018.

\bibitem{Lam:2017ta}
M.~Lam, B.~Mahasseni, and S.~Todorovic.
\newblock {Fine-grained recognition as hsnet search for informative image
  parts}.
\newblock In {\em CVPR}, 2017.

\bibitem{Lenc:2016tz}
K.~Lenc and A.~Vedaldi.
\newblock {Learning covariant feature detectors}.
\newblock In {\em ECCV Workshops}, 2016.

\bibitem{Li2018analogy}
Z.~Li, Y.~Tang, and Y.~He.
\newblock Unsupervised disentangled representation learning with analogical
  relations.
\newblock In {\em IJCAI}, 2018.

\bibitem{Lim:2018uo}
J.~Lim, Y.~Yoo, B.~Heo, and J.~Y. Choi.
\newblock {Pose transforming network: Learning to disentangle human posture in
  variational auto-encoded latent space}.
\newblock {\em Pattern Recognit. Lett.}, 2018.

\bibitem{Liu:2016vv}
Z.~Liu, P.~Luo, S.~Qiu, X.~Wang, and X.~Tang.
\newblock {Deepfashion: Powering robust clothes recognition and retrieval with
  rich annotations}.
\newblock In {\em CVPR}, 2016.

\bibitem{Liu:2015vj}
Z.~Liu, P.~Luo, X.~Wang, and X.~Tang.
\newblock {Deep learning face attributes in the wild}.
\newblock In {\em ICCV}, 2015.

\bibitem{Liu:2016td}
Z.~Liu, S.~Yan, P.~Luo, X.~Wang, and X.~Tang.
\newblock {Fashion landmark detection in the wild}.
\newblock In {\em ECCV}, 2016.

\bibitem{Ma:2017uu}
L.~Ma, X.~Jia, Q.~Sun, B.~Schiele, T.~Tuytelaars, and L.~V. Gool.
\newblock {Pose guided person image generation}.
\newblock In {\em NIPS}, 2017.

\bibitem{Ma:2017wq}
L.~Ma, Q.~Sun, S.~Georgoulis, L.~V. Gool, B.~Schiele, and M.~Fritz.
\newblock {Disentangled person image generation}.
\newblock {\em CVPR}, 2017.

\bibitem{Mesnil:2013hi}
G.~Mesnil, A.~Bordes, J.~Weston, G.~Chechik, and Y.~Bengio.
\newblock {Learning semantic representations of objects and their parts}.
\newblock {\em Mach Learn}, 2013.

\bibitem{Newell:2016vq}
A.~Newell, K.~Yang, and J.~Deng.
\newblock {Stacked hourglass networks for human pose estimation}.
\newblock {\em ECCV}, 2016.

\bibitem{Nguyen:2013vk}
T.~D. Nguyen, T.~Tran, D.~Q. Phung, and S.~Venkatesh.
\newblock {Learning parts-based representations with nonnegative restricted
  boltzmann machine}.
\newblock In {\em ACML}, 2013.

\bibitem{Novotny:2017ta}
D.~Novotny, D.~Larlus, and A.~Vedaldi.
\newblock {Anchornet: A weakly supervised network to learn geometry-sensitive
  features for semantic matching}.
\newblock In {\em CVPR}, 2017.

\bibitem{Pedersoli:2014ta}
M.~Pedersoli, R.~Timofte, T.~Tuytelaars, and L.~J.~V. Gool.
\newblock {Using a deformation field model for localizing faces and facial
  points under weak supervision}.
\newblock In {\em CVPR}, 2014.

\bibitem{Pfister:2015uo}
T.~Pfister, J.~Charles, and A.~Zisserman.
\newblock {Flowing convnets for human pose estimation in videos}.
\newblock In {\em ICCV}, 2015.

\bibitem{Ranjan:2016vv}
R.~Ranjan, V.~M. Patel, and R.~Chellappa.
\newblock {Hyperface: A deep multi-task learning framework for face detection,
  landmark localization, pose estimation, and gender recognition}.
\newblock {\em TPAMI}, 2017.

\bibitem{Ronneberger:2015gk}
O.~Ronneberger, P.~Fischer, and T.~Brox.
\newblock {U-net: Convolutional networks for biomedical image segmentation}.
\newblock In {\em MICCAI}, 2015.

\bibitem{Ross:2006uc}
D.~A. Ross and R.~S. Zemel.
\newblock {Learning parts-based representations of data}.
\newblock {\em JMLR}, 2006.

\bibitem{Russakovsky2015imagenet}
O.~Russakovsky, J.~Deng, H.~Su, J.~Krause, S.~Satheesh, S.~Ma, Z.~Huang,
  A.~Karpathy, A.~Khosla, M.~Bernstein, et~al.
\newblock Imagenet large scale visual recognition challenge.
\newblock {\em ICCV}, 2015.

\bibitem{Shu:2018ua}
Z.~Shu, M.~Sahasrabudhe, R.~A. G{\"u}ler, D.~Samaras, N.~Paragios, and
  I.~Kokkinos.
\newblock {Deforming autoencoders: Unsupervised disentangling of shape and
  appearance}.
\newblock In {\em ECCV}, 2018.

\bibitem{Siarohin:2018wk}
A.~Siarohin, E.~Sangineto, S.~Lathuili{\`e}re, and N.~Sebe.
\newblock {Deformable gans for pose-based human image generation}.
\newblock {\em CVPR}, 2018.

\bibitem{Singh:2012un}
S.~Singh, A.~Gupta, and A.~A. Efros.
\newblock {Unsupervised discovery of mid-level discriminative patches}.
\newblock In {\em ECCV}, 2012.

\bibitem{Szegedy2015inception}
C.~Szegedy, W.~Liu, Y.~Jia, P.~Sermanet, S.~Reed, D.~Anguelov, D.~Erhan,
  V.~Vanhoucke, and A.~Rabinovich.
\newblock Going deeper with convolutions.
\newblock In {\em CVPR}, 2015.

\bibitem{Thewlis:2017wg}
J.~Thewlis, H.~Bilen, and A.~Vedaldi.
\newblock {Unsupervised learning of object frames by dense equivariant image
  labelling}.
\newblock In {\em NIPS}, 2017.

\bibitem{Thewlis:2017wi}
J.~Thewlis, H.~Bilen, and A.~Vedaldi.
\newblock {Unsupervised learning of object landmarks by factorized spatial
  embeddings}.
\newblock In {\em ICCV}, 2017.

\bibitem{Toshev:2014tp}
A.~Toshev and C.~Szegedy.
\newblock {Deeppose: Human pose estimation via deep neural networks}.
\newblock In {\em CVPR}, 2014.

\bibitem{ufer2017deep}
N.~Ufer and B.~Ommer.
\newblock Deep semantic feature matching.
\newblock In {\em Proceedings of the IEEE Conference on Computer Vision and
  Pattern Recognition}, pages 6914--6923, 2017.

\bibitem{Wah:2011vq}
C.~Wah, S.~Branson, P.~Welinder, P.~Perona, and S.~Belongie.
\newblock {The caltech-ucsd birds-200-2011 dataset}.
\newblock Technical report, California Institute of Technology, 2011.

\bibitem{Wei:2016ws}
S.-E. Wei, V.~Ramakrishna, T.~Kanade, and Y.~Sheikh.
\newblock {Convolutional pose machines}.
\newblock In {\em CVPR}, 2016.

\bibitem{Wu:2017vc}
Y.~Wu and Q.~Ji.
\newblock {Robust facial landmark detection under significant head poses and
  occlusion}.
\newblock {\em CVPR}, 2015.

\bibitem{Xiao:2017}
T.~Xiao, S.~Li, B.~Wang, L.~Lin, and X.~Wang.
\newblock Joint detection and identification feature learning for person
  search.
\newblock In {\em CVPR}. IEEE, 2017.

\bibitem{Xing:2018un}
X.~Xing, R.~Gao, T.~Han, S.-C. Zhu, and Y.~N. Wu.
\newblock {Deformable generator network: Unsupervised disentanglement of
  appearance and geometry}.
\newblock {\em arXiv preprint arXiv:1806.06298}, 2018.

\bibitem{Yang:2016uo}
W.~Yang, W.~Ouyang, H.~Li, and X.~Wang.
\newblock {End-to-end learning of deformable mixture of parts and deep
  convolutional neural networks for human pose estimation}.
\newblock In {\em CVPR}, 2016.

\bibitem{Yildirim:2015ur}
I.~Yildirim, T.~D. Kulkarni, W.~Freiwald, and J.~B. Tenenbaum.
\newblock {Efficient analysis-by-synthesis in vision: A computational
  framework, behavioral tests, and modeling neuronal representations}.
\newblock In {\em CogSci}, 2015.

\bibitem{Yu:2016vi}
X.~Yu, F.~Zhou, and M.~Chandraker.
\newblock {Deep deformation network for object landmark localization}.
\newblock In {\em ECCV}, 2016.

\bibitem{Zhang:2008uj}
W.~Zhang, J.~Sun, and X.~Tang.
\newblock {Cat head detection - how to effectively exploit shape and texture
  features}.
\newblock In {\em ECCV}, 2008.

\bibitem{Zhang:2013tr}
W.~Zhang, M.~Zhu, and K.~G. Derpanis.
\newblock {From actemes to action: A strongly-supervised representation for
  detailed action understanding}.
\newblock In {\em ICCV}, 2013.

\bibitem{Zhang:2018vz}
Y.~Zhang, Y.~Guo, Y.~Jin, Y.~Luo, Z.~He, and H.~Lee.
\newblock Unsupervised discovery of object landmarks as structural
  representations.
\newblock In {\em CVPR}, 2018.

\bibitem{Zhang:2014wy}
Z.~Zhang, P.~Luo, C.~C. Loy, and X.~Tang.
\newblock {Facial landmark detection by deep multi-task learning}.
\newblock In {\em ECCV}, 2014.

\bibitem{Zhang:2016vx}
Z.~Zhang, P.~Luo, C.~C. Loy, and X.~Tang.
\newblock {Learning deep representation for face alignment with auxiliary
  attributes}.
\newblock {\em TPAMI}, 2016.

\bibitem{Zhu:2015tz}
S.~Zhu, C.~Li, C.~C. Loy, and X.~Tang.
\newblock {Face alignment by coarse-to-fine shape searching}.
\newblock In {\em CVPR}, 2015.

\end{thebibliography}
}
\clearpage
\appendix
\onecolumn
\begin{multicols}{2}
\section{Supplementary Material}
\subsection{Disentangled Representation}
\textbf{Local Appearance Transfer.}
In Fig.~\ref{fig:partswaps}, we show  results for successively swapping part appearance on the Deep Fashion dataset as an extension to Fig. 8 in the main paper.
\end{multicols}
\begin{figure*}[h!]
    \vspace*{-1.em}
	\centering
	\begin{subfigure}{.42\linewidth}
	\includegraphics[trim={0cm 0cm 0cm 0cm},clip, width=1.\linewidth]{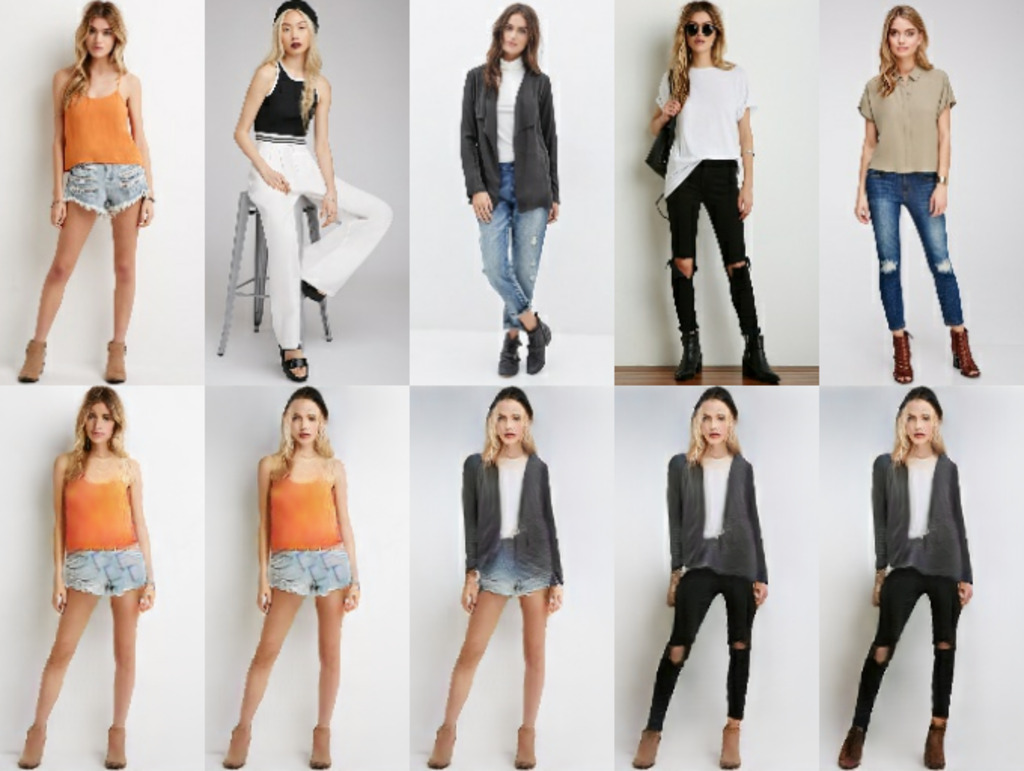}
	\label{fig:part3_21}
	\end{subfigure}\hspace{0.03\textwidth}
	\vspace*{-1.em}
	\centering
	\begin{subfigure}{.42\linewidth}
	\includegraphics[trim={0cm 0cm 0cm 0cm},clip, width=1.\linewidth]{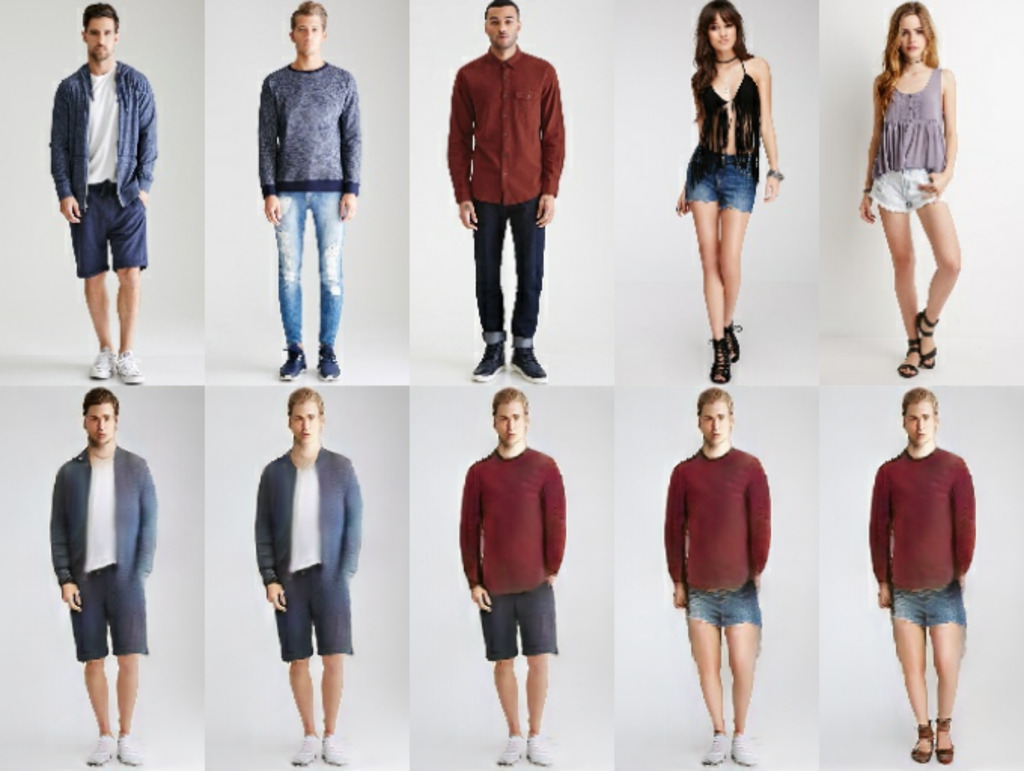}
	\label{fig:part3_30}
	\end{subfigure}
	\vspace*{-1.em}
	\centering
	\begin{subfigure}{.42\linewidth}
	\centering
	\includegraphics[trim={0cm 0cm 0cm 0cm},clip, width=1.\linewidth]{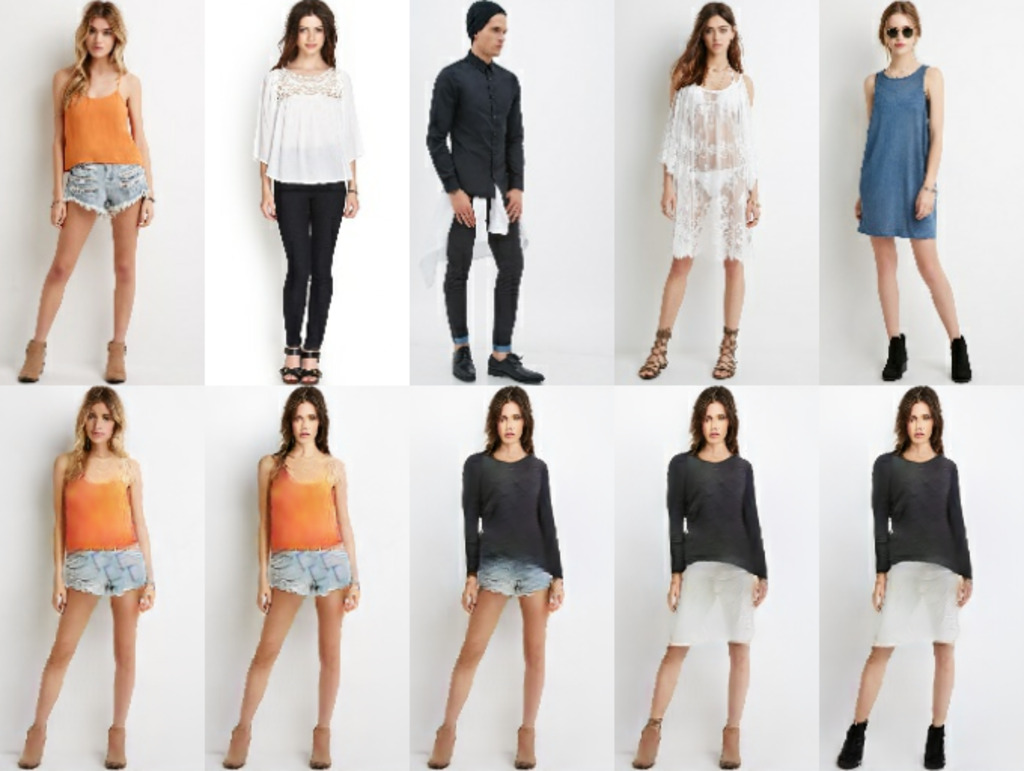}
	\label{fig:part3_30}
	\end{subfigure}\hspace{0.03\textwidth}
	\label{fig:partswaps2}
	\centering
	\begin{subfigure}{.42\linewidth}
	\centering
	\includegraphics[trim={0cm 0cm 0cm 0cm},clip, width=1.\linewidth]{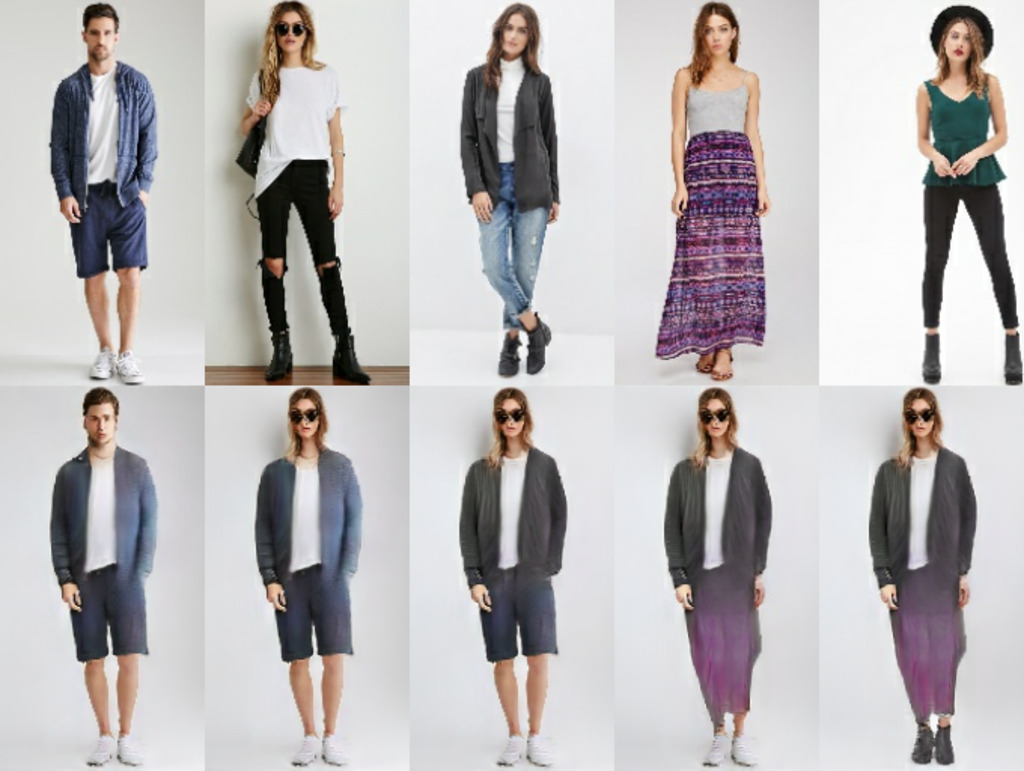}
	\label{fig:part3_21}
	\end{subfigure}
	\centering
	\begin{subfigure}{.42\linewidth}
	\centering
	\includegraphics[trim={0cm 0cm 0cm 0cm},clip, width=1.\linewidth]{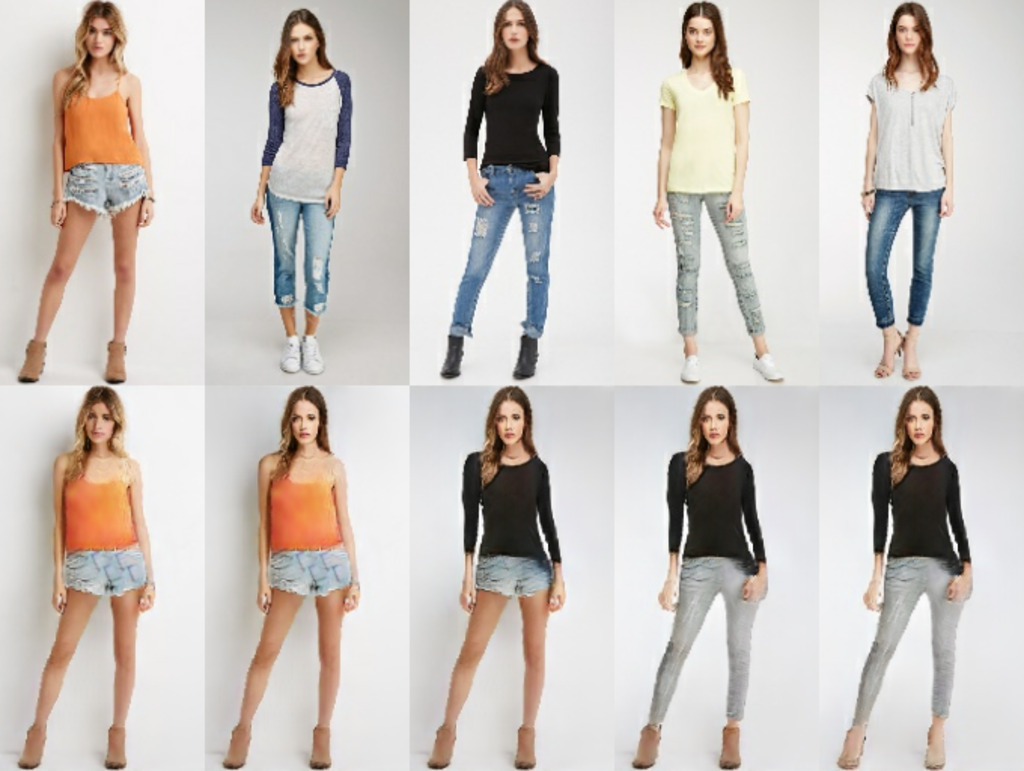}
	\label{fig:part3_30}
	\end{subfigure}\hspace{0.03\textwidth}
	\centering
	\begin{subfigure}{.42\linewidth}
	\centering
	\includegraphics[trim={0cm 0cm 0cm 0cm},clip, width=1.\linewidth]{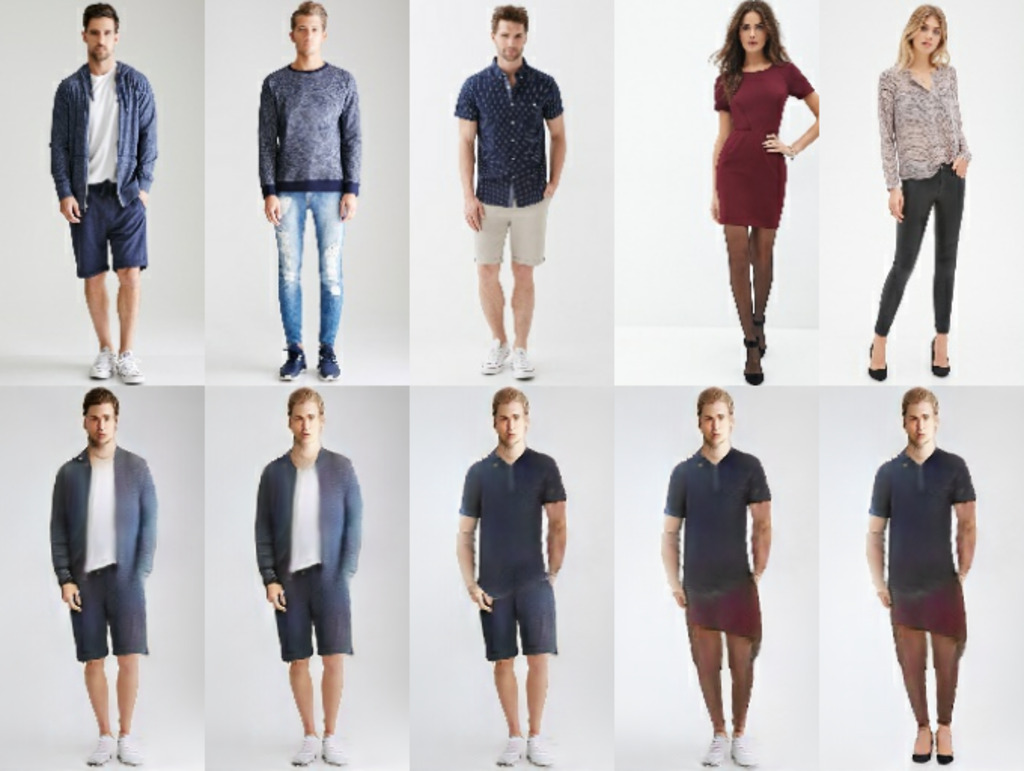}
	\label{fig:part3_30}
	\end{subfigure}
	\caption{
	Successively altering the appearance of individual parts. We show 6 examples of successively altering appearances of parts using different source images. In each example we start from the original appearance (left-most column). The top row shows ground-truth images (taken from the test-set), which act as the source for the part appearance to be altered. The bottom row then illustrates the new synthesized image, which is generated based on the already altered part appearances plus the current appearance modification. Part appearances are altered in fixed order: head, upper body, legs, feet. 
	}
	\label{fig:partswaps}
\end{figure*}

\begin{figure*}[h!]
	\centering
	\begin{subfigure}{.45\linewidth}
	\centering
	\includegraphics[trim={0cm 0cm 0cm 0cm},clip, width=1.\linewidth]{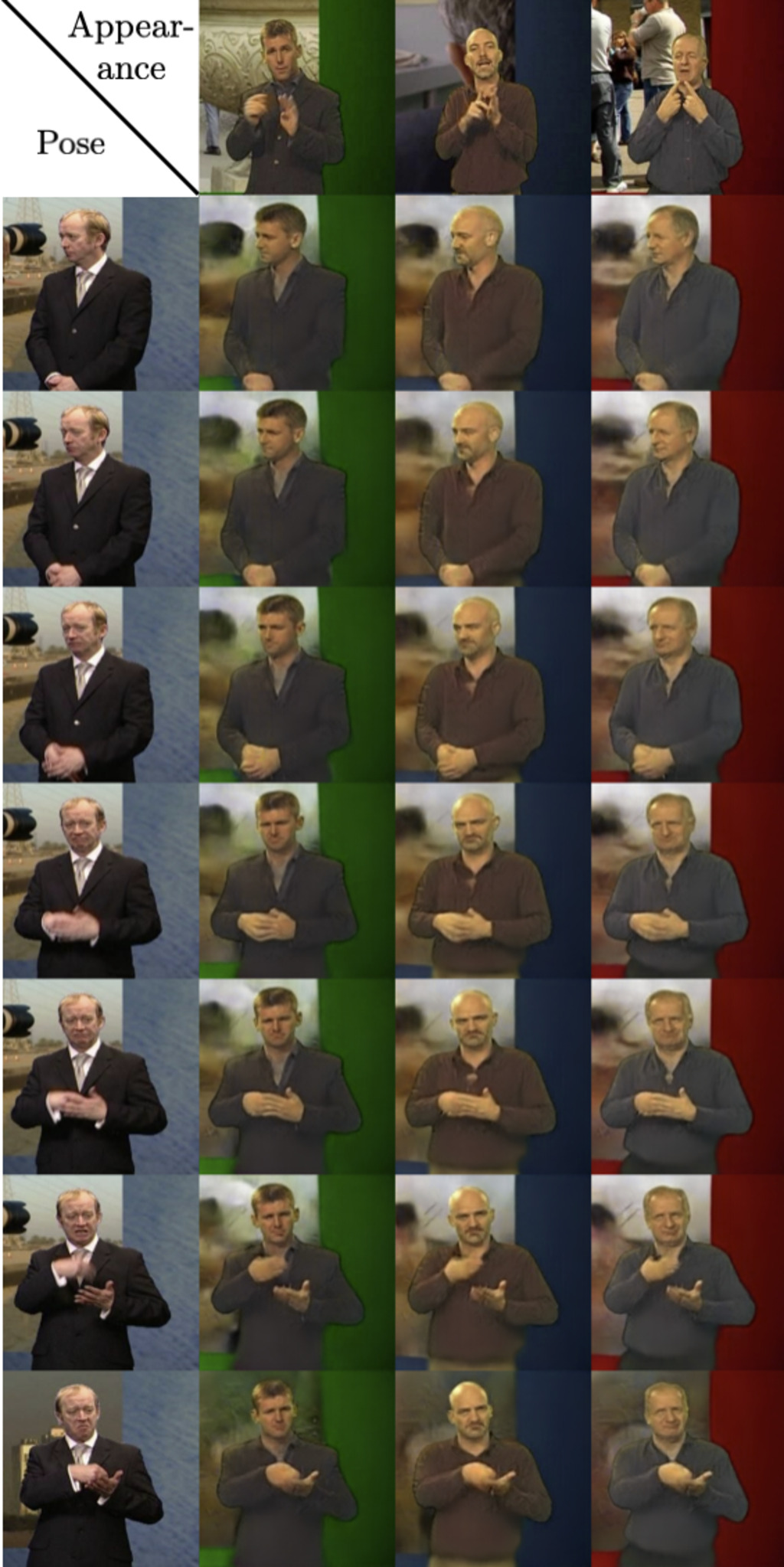}
	\end{subfigure}\hspace{0.03\textwidth}
	\centering
	\begin{subfigure}{.45\linewidth}
	\centering
	\includegraphics[trim={0cm 0cm 0cm 0cm},clip, width=1.\linewidth]{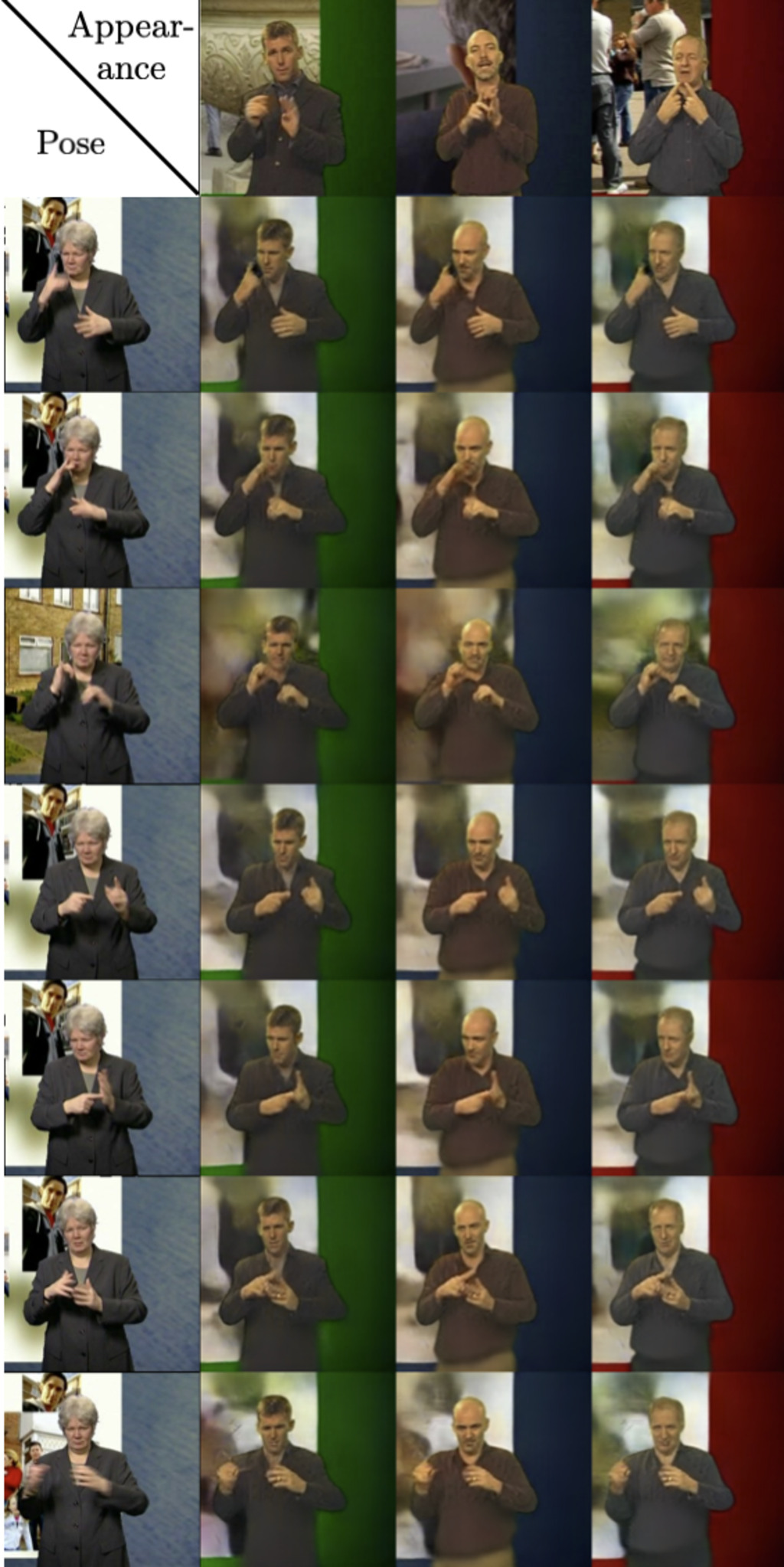}
	\end{subfigure}
	\caption{Generated sequence on BBC Pose from a target pose sequence (leftmost column) and target appearances (top row). }\label{fig:bbc1}
\end{figure*}

\clearpage
\begin{multicols}{2}
\textbf{Video-to-Video Translation.}
In Fig.~\ref{fig:bbc1} we show sequences of a frame-to-frame appearance-shape transfer on the BBC Pose dataset, as an extension to main paper, Fig. 7.
\end{multicols}
\clearpage

\twocolumn
\subsection{Landmark Discovery}
\begin{figure}
    \centering
    \includegraphics[trim={0cm 0cm 0cm 0cm},clip, width=.9\linewidth]{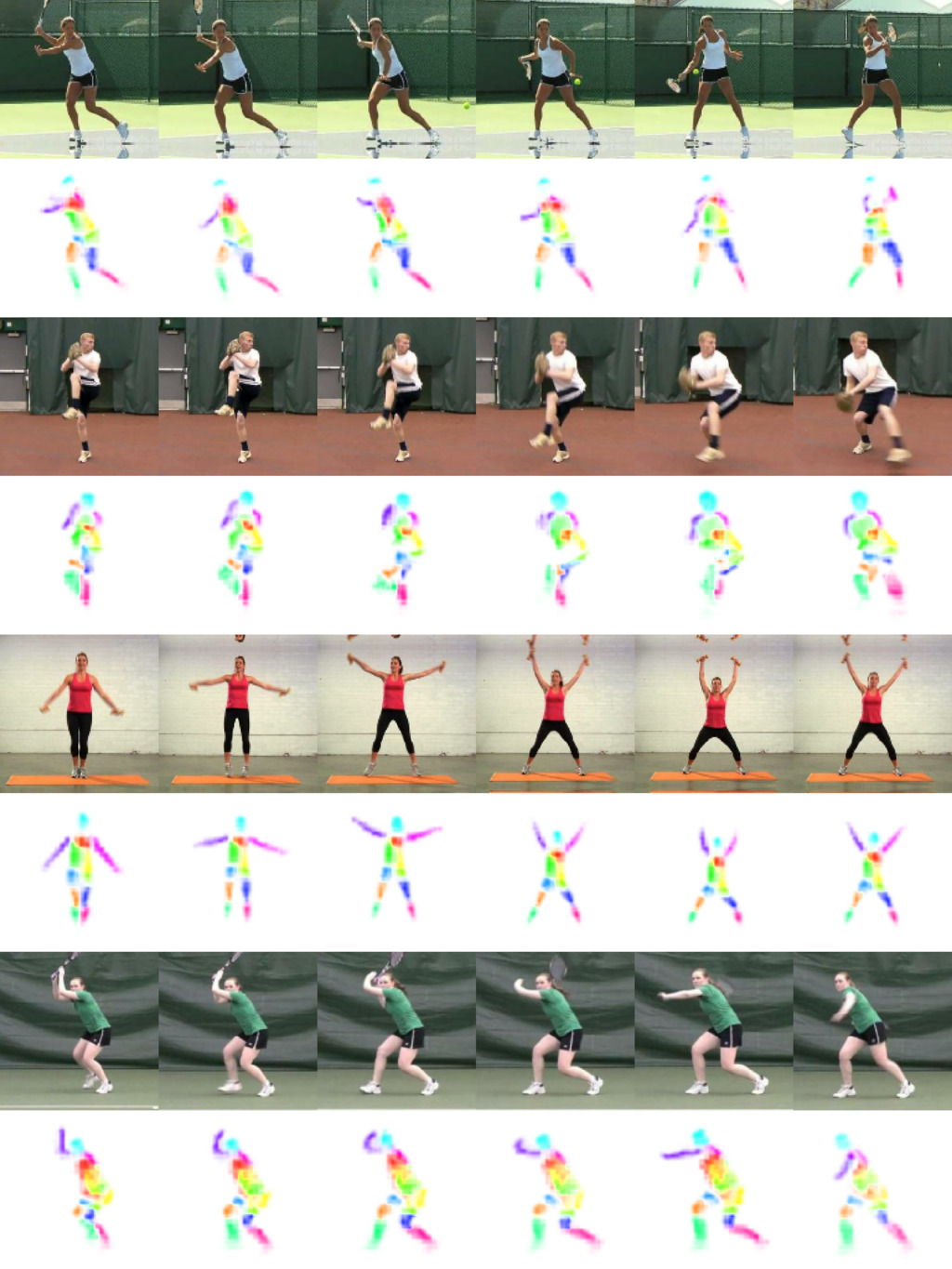}
    \caption{Showing 12 out of 16 part activation maps on Penn Action.}
    \label{fig:shape}
\end{figure}
\begin{figure}
    \centering
    \includegraphics[trim={0cm 0cm 0cm 0cm},clip, width=.9\linewidth]{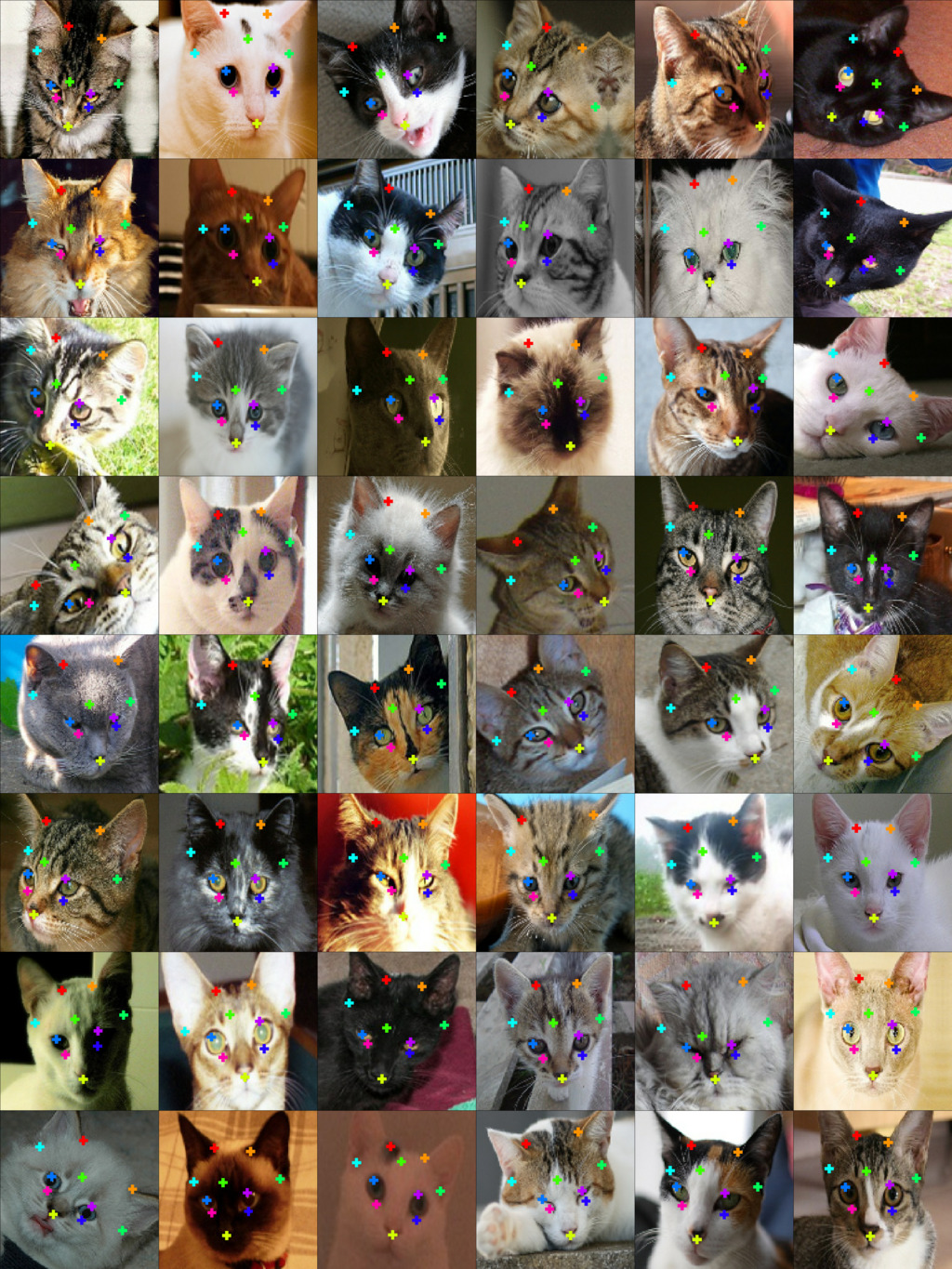}
    \caption{Discovering 10 landmarks on Cat Head.}
    \label{fig:kp_cats}
\end{figure}
\begin{figure}
    \centering
    \includegraphics[trim={0cm 0cm 0cm 0cm},clip, width=.9\linewidth]{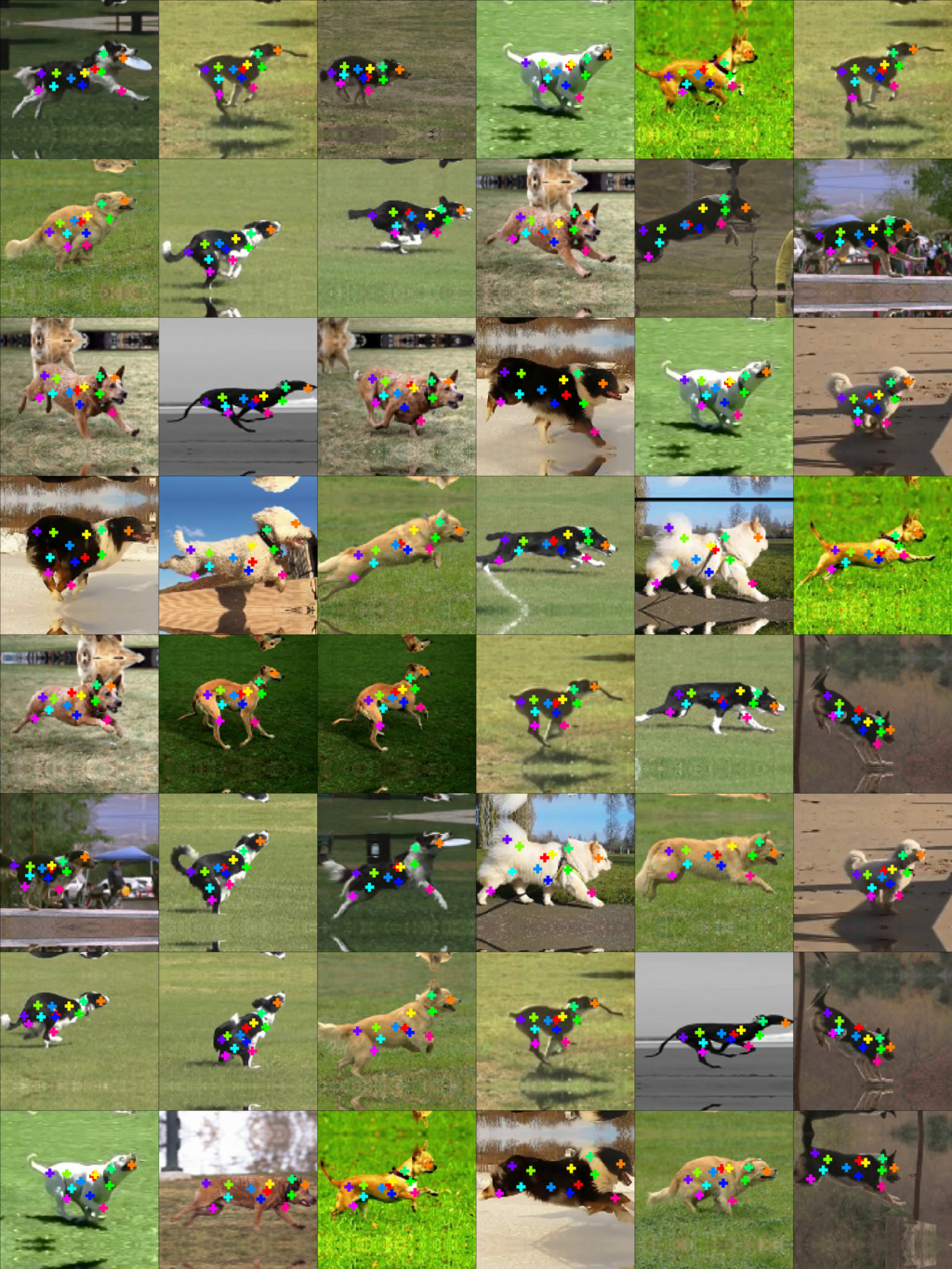}
    \caption{Discovering 10 landmarks on Dogs Run.}
    \label{fig:kp_dogs}
\end{figure}
\begin{figure}
    \centering
    \includegraphics[trim={0cm 0cm 0cm 0cm},clip, width=0.9\linewidth]{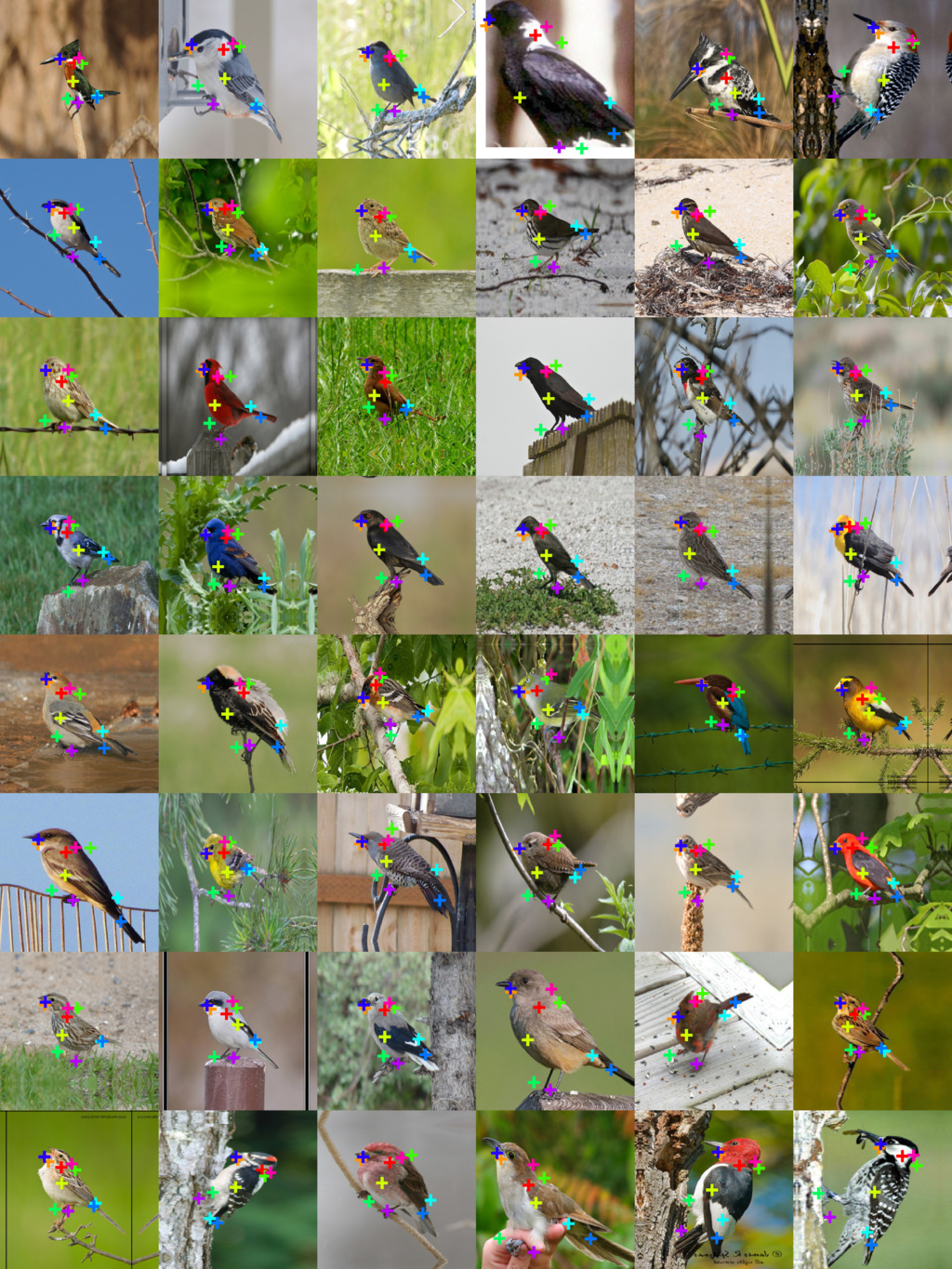}
    \caption{Discovering 10 landmarks on CUB-200-2011}
    \label{fig:kp_birds}
\end{figure}
\begin{figure}
    \centering
    \includegraphics[trim={0cm 0cm 0cm 0cm},clip, width=0.9\linewidth]{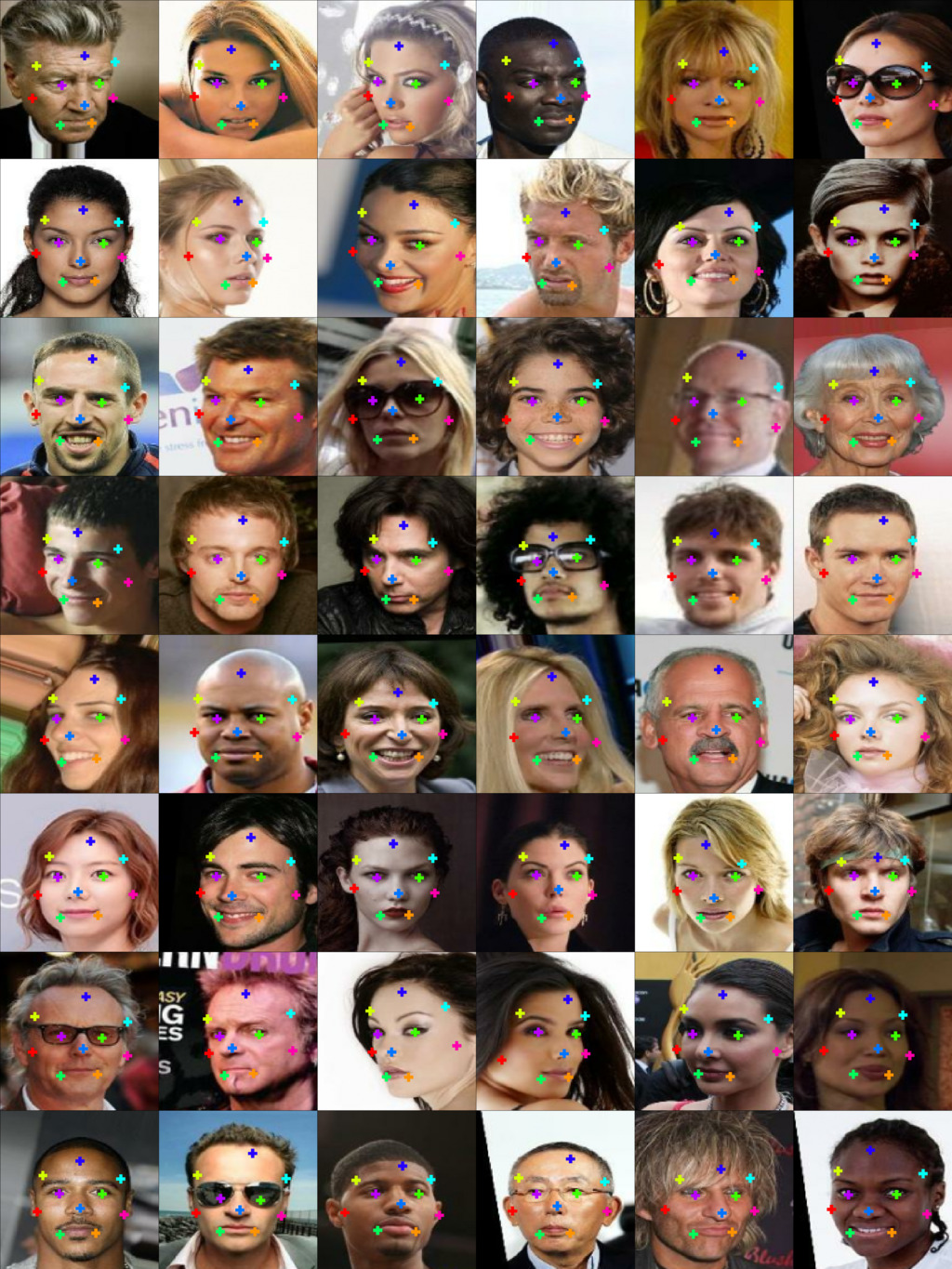}
    \caption{Discovering 10 landmarks on CelebA.}
    \label{fig:kp_celeb}
\end{figure}
\begin{figure}
    \centering
    \includegraphics[trim={0cm 0cm 0cm 0cm},clip, width=0.9\linewidth]{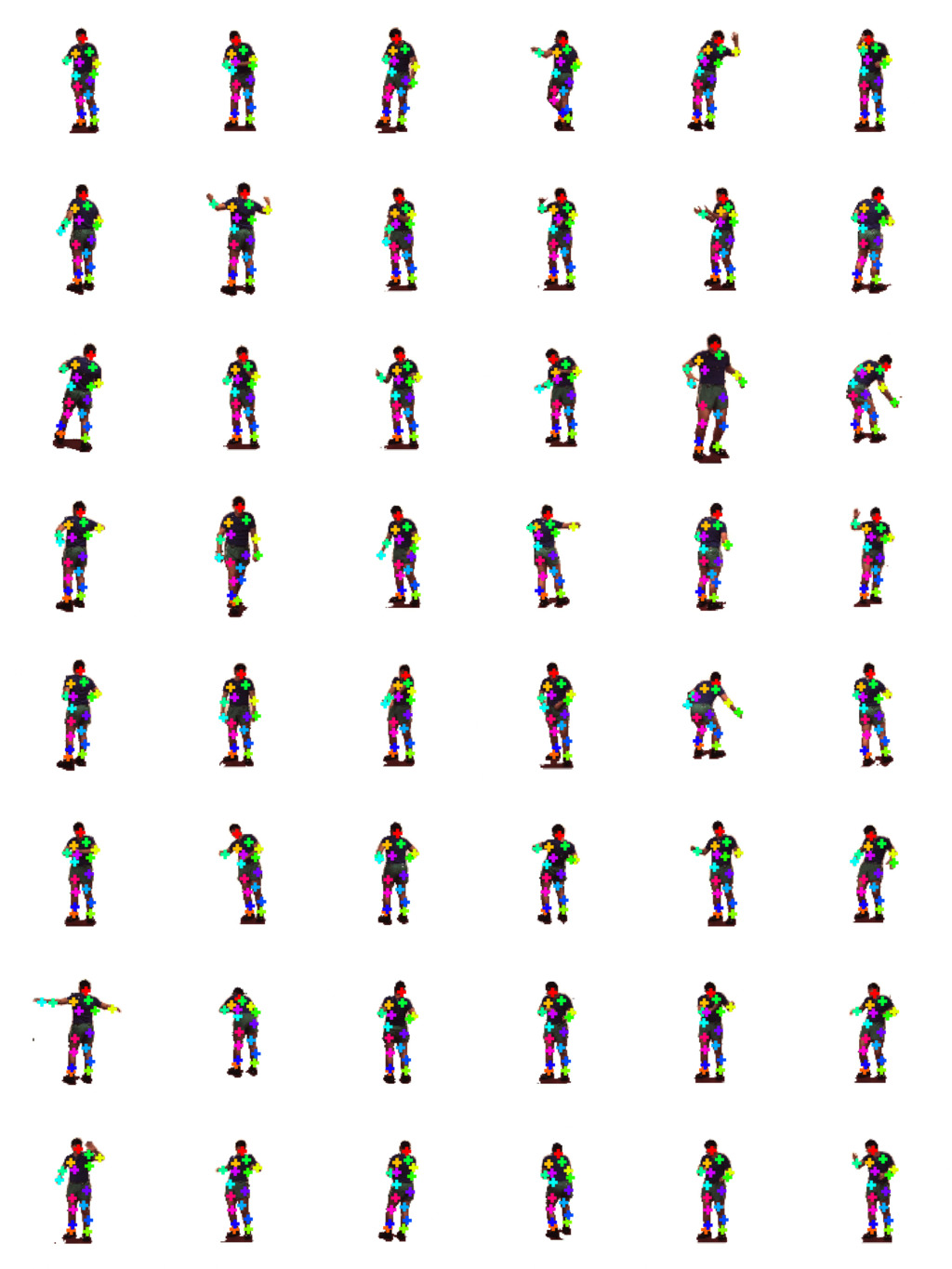}
    \caption{Discovering 10 landmarks on Human3.6M.}
    \label{fig:kp_human}%
\end{figure}
\begin{figure}
    \centering
    \includegraphics[trim={0cm 0cm 0cm 0cm},clip, width=0.9\linewidth]{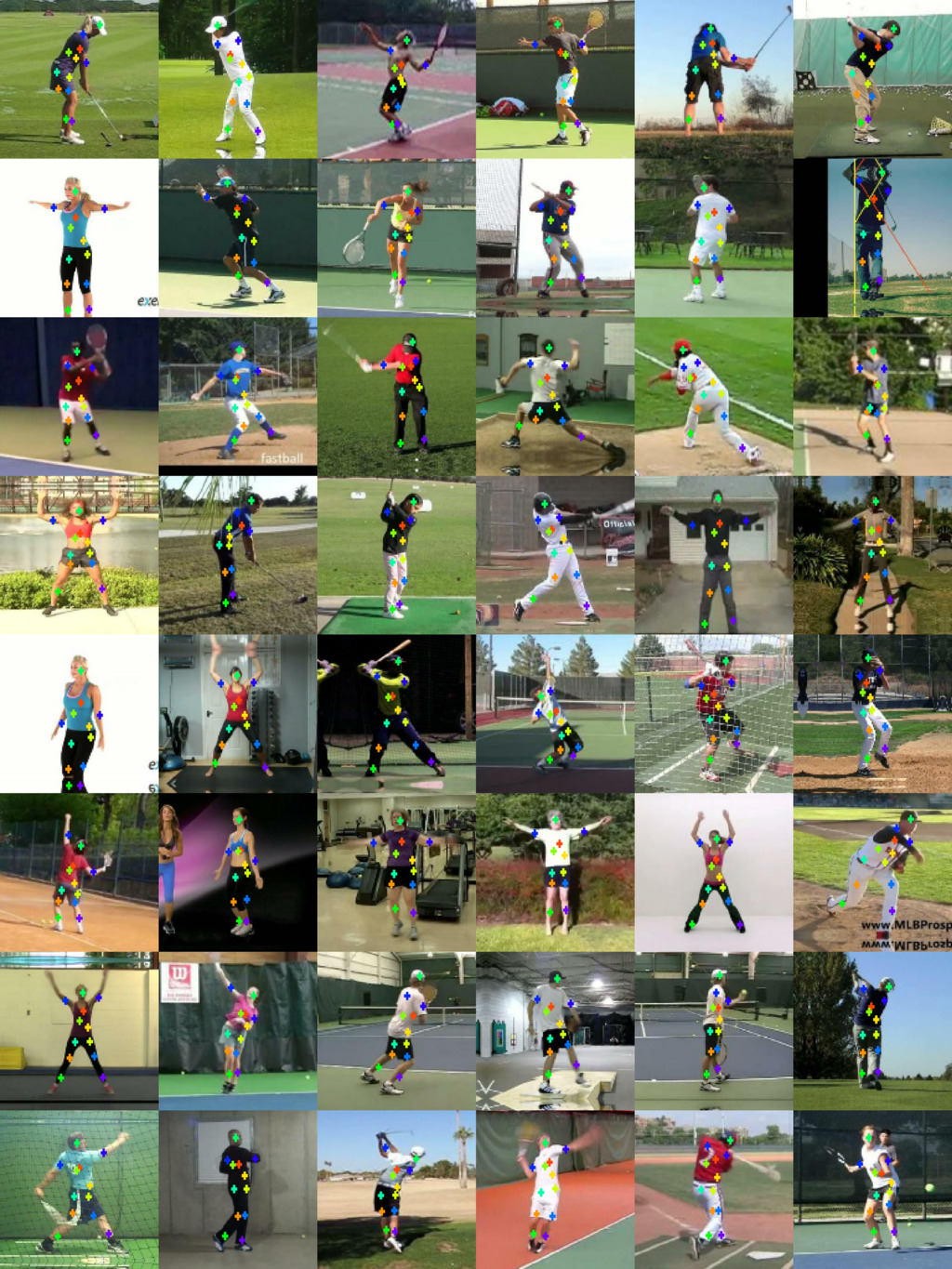}
    \caption{Showing 12 out of 16 landmarks on Penn Action.}
    \label{fig:kp_penn}
\end{figure}

\textbf{Part Activations.}
In Fig. \ref{fig:shape} we show part activation maps on video sequences from the Penn Action dataset, as an extension to main paper, Fig. 3.\\
\textbf{Landmark Discovery.}
We present unsupervised landmark discovery results on the following datasets: Cat Head (Fig. \ref{fig:kp_cats}), Dogs Run (Fig. \ref{fig:kp_dogs}), CUB-200-2011 (Fig. \ref{fig:kp_birds}), CelebA (Fig. \ref{fig:kp_celeb}), Human3.6M (Fig. \ref{fig:kp_human}) and Penn Action (Fig. \ref{fig:kp_penn}), as an extension to main paper, Fig. 4.

\clearpage

\clearpage
\subsection{Implementation Details and Settings}

\textbf{Implementation Details.}
Table \ref{tab:hypers} gives an overview over the different settings for the datasets we used in our experiments.

The architecture of $E_{\sigma}$ and $E_{\alpha}$ is based on the implementation \cite{Walidtf} of the stacked hourglass architecture \cite{Newell:2016vq}. In a first step the image with input resolution $h \times w \times 3$ is processed by a series of convolutions to image features of dimension $64 \times 64 \times 256$. The hourglass modules of $E_{\sigma}$ and $E_{\alpha}$ operate on a maximal resolution of $64 \times 64$, thus part activation maps and the localized image appearance encoding both have a spatial dimension of $64 \times 64$.  $E_{\sigma}$ reaches its lowest resolution at $4 \times 4$ pixels whereas $E_{\alpha}$ has its lowest resolution at $32 \times 32$ pixels. All residual blocks of the hourglass modules have $256$ feature channels. The decoder is a variant of a U-Net \cite{Ronneberger:2015gk} operating at a resolution of $h \times w$ pixels. Different from a standard U-Net we do not learn the downsampling stream. Through skip connections the approximate part activations maps are passed to the upsampling stream with the appropriate resolutions. We distribute the local appearance encoding together with the corresponding approximate part activation maps into a multiscale bottleneck of resolution $4 \times 4$ to $16 \times 16$ in the Unet. The convolutional filters in the first up-sampling stage of the U-Net have $512$ feature channels. The number of feature channels is halved every two upsampling stages.
\\
The $\ell_1$ reconstruction loss (Eq. 2) is weighted locally around the part activations $\sigma_i(x)$. For this, we multiply the loss with a soft mask. 
For an image $x$ at pixel $u$ the mask takes the form 
$\textrm{mask}[u] = \text{min}\big(\sum_i  \frac{1}{1 + \lVert u -  \mu[\sigma_i(x)]/\lambda_\text{scal} \rVert}, 1\big)$ where $\lambda_\text{scal}$ is a hyperparameter. \\
% We do not propagate gradients through the means $\mu([\sigma_i(x)])$ of the mask.\\
\textbf{Decoder Approximation}
The decoder receives approximated part activation maps 
$\tilde{\sigma}_i(a(x))$ (cf. Eq. 7). 
We utilize two variants for this approximation: \emph{i)} $\Sigma_i$ is fixed to the identity matrix 
\emph{ii)} $\Sigma_i$ is the covariance of $\sigma_i(a(x))/\sum_{u \in \Lambda} \sigma_i(a(x))[u]$. In practice \emph{i}) leads to more confined part shapes and is used for experiments involving keypoint regression.

\textbf{Adversarial Task.}
To improve the quality of image generations, we implement a variant of the adversarial task, as presented in \cite{Isola2017image}: A discriminator is trained to classify $N \times N$ image patches as real or fake.
Using the mean locations of part shapes as center points, we extract image patches of size $49 \times 49$ from the real image $x$ and the generated image $\hat{x}$. As conditioning, the discriminator is additionally provided with corresponding patches extracted on the stack of approximated part activations $\tilde{\sigma}_i(x)$. The discriminator is implemented as a lightweight CNN architecture consisting of 4 convolution layers with stride 2 followed by a dense layer.
The adversarial task is trained simultaneously with the main objective function (Eq. 4), no subsequent fine-tuning step is necessary.

\begin{table}[H]
	\centering
	\begin{tabular}{l|cccc}
		\hline
		\thead{Dataset} &  \thead{parts}&  \thead{resolution} & \thead{lr.}  &  \thead{advers.} \\ \hline
		\thead{Cat Head \cite{Zhang:2008uj}} & \thead{10\ /\ 20} & \thead{$128\times128$} &  \thead{0.001}   &  \thead{\ding{55}}\\
		\thead{CelebA \cite{Liu:2015vj}}  & \thead{10}  & \thead{$128\times128$}  &\thead{0.001}  &  \thead{\ding{55}}\\
		\thead{Human3.6M \cite{Ionescu:2014ua}}& \thead{16}  & \thead{$128\times128$}  &  \thead{0.0002}  &  \thead{\ding{55}}  \\
		\thead{Penn Action \cite{Zhang:2013tr}} & \thead{16} &  \thead{$128\times128$}  & \thead{0.0002}  &   \thead{\ding{55}}\\
		\thead{Dogs Run (own)} & \thead{12} &   \thead{$128\times128$}  & \thead{0.001} &  \thead{\ding{55}} \\
		\thead{CUB-200-2011 \cite{Wah:2011vq}}& \thead{10} &   \thead{$128\times128$} & \thead{0.001}   &   \thead{\ding{55}}\\
		\thead{BBC Pose Regression \cite{Charles:2013tb}} & \thead{30} &  \thead{$128\times128$} & \thead{0.001}   &  \thead{\ding{55}} \\
		\thead{BBC Pose Synthesis \cite{Charles:2013tb}}& \thead{40} &  \thead{$256\times256$} & \thead{0.001}   &  \thead{\ding{51}}\\
		\thead{Deep Fashion  \cite{Liu:2016vv, Liu:2016td}} & \thead{16} &  \thead{$256\times256$} & \thead{0.001}  &  \thead{\ding{51}}  \\
	\end{tabular}
		\centering
	\caption{Settings for different experiments: number of landmarks, input resolution, learning rate of Adam optimizer, adversarial task }
	\label{tab:hypers}
\end{table}

\end{document}